\documentclass[10pt,journal,compsoc]{IEEEtran}

\usepackage[noadjust]{cite}
\usepackage[dvipsnames]{xcolor}
\usepackage{tikz}
\usetikzlibrary{shapes,calc,matrix}
\usepackage[british]{babel}
\usepackage[utf8]{inputenc}
\usepackage[normalem]{ulem}
\usepackage[T1]{fontenc}
\usepackage{mathtools}

\usepackage{tikz}
\usetikzlibrary{matrix,chains,positioning,decorations.pathreplacing,arrows}
\usetikzlibrary{arrows.meta, chains, positioning, quotes}
\usetikzlibrary{positioning,calc}

\usepackage{url}
\usepackage{amsfonts}
\usepackage{amssymb}
\usepackage{graphicx}
\usepackage{caption} 
\usepackage{subfig}
\usepackage{enumitem}
\usepackage{bm}
\usepackage{stfloats}
\usepackage[final]{pdfpages}
\usepackage{slashbox}
\usepackage[normalem]{ulem}
\usepackage{systeme}
\usepackage{algorithm, float}
\usepackage{algorithmic} 
\usepackage[squaren, Gray]{SIunits}
\usepackage[acronym,toc,nonumberlist, nopostdot, style=super, nogroupskip]{glossaries}
\usepackage{hyperref}
\usepackage{amsmath}
\usepackage[thmmarks,amsmath]{ntheorem}
\usepackage[figurename=Fig.]{caption}

\DeclareMathOperator*{\argreduce}{arg\,reduce}
\DeclareMathOperator{\Tr}{Tr}
\makeatletter
\renewcommand\p@subfigure{\thefigure\,}
\makeatother
\newcommand{\rulesep}{\unskip\ \vrule\ }

\tikzset{%
  every neuron/.style={
    circle,
    draw,
    minimum size=1cm
  },
  neuron missing/.style={
    draw=none, 
    scale=4,
    text height=0.25cm,
    execute at begin node=\color{black}$\vdots$
  },
}

\definecolor{brightpink}{rgb}{1.0, 0.0, 0.5}

\begin{document}

%

\title{Deep Matrix Factorizations}

\author{Pierre De Handschutter \quad 
Nicolas Gillis \quad Xavier Siebert\thanks{This work was supported by the European Research Council (ERC starting grant n$^\text{o}$ 679515), and by the Fonds de la Recherche Scientifique - FNRS and the Fonds Wetenschappelijk Onderzoek - Vlaanderen (FWO) under EOS Project no O005318F-RG47. 
}
\thanks{
The authors are with the Department of Mathematics and Operational Research, Faculté Polytechnique, Université de Mons, 7000 Mons, Belgium. Pierre De Handschutter is a research fellow of the Fonds de la Recherche Scientifique - FNRS. 
E-mails: \{pierre.dehandschutter,
nicolas.gillis,
xavier.siebert\}@umons.ac.be.
}}
\hyphenation{Weakly-supervised}


%


\maketitle

\begin{abstract}
Constrained low-rank matrix approximations 
have been known for decades as powerful linear dimensionality reduction techniques to be able to extract the information contained in large data sets in a relevant way. However, such low-rank approaches are unable to mine complex, interleaved features that underlie hierarchical semantics. Recently, deep matrix factorization (deep MF) was introduced to deal with the extraction of several layers of features and has been shown to reach outstanding performances on unsupervised tasks. 
Deep MF was motivated by the success of deep learning, as it is conceptually close to some neural networks paradigms. In this paper, we present the main models, algorithms, and applications of deep MF through a comprehensive literature review. We also discuss theoretical questions and perspectives of research. 
\end{abstract}
%
\IEEEpeerreviewmaketitle
\section{Introduction} \label{sec:intro}

In the current era of data deluge, the automatic extraction of interpretable features in unlabelled data sets is a key challenge. For many years, linear algebra tools have been used to deal with such tasks. Among these techniques, the constrained low-rank matrix approximations (CLRMA)~\cite{udell2016generalized}  mine relevant information from large data sets and have therefore been drawing attention of numerous researchers.  In practice, many data sets appear to be well approximated by low-rank matrices~\cite{udell2019big}, and hence CLRMA are particularly appropriate to extract pertinent information. Within this general framework, some well-known techniques such as principal component analysis (PCA)~\cite{wold1987principal}, singular values decomposition (SVD)~\cite{golub1971singular}, 
sparse coding~\cite{papyan2018theoretical}, 
sparse component analysis (SCA)~\cite{georgiev2005sparse}, and 
non-negative matrix factorization (NMF)~\cite{lee1999learning}, to name only a few, have been used in many applications for the last decades. These variants mostly differ by the function chosen to measure the quality of the approximation and by the additional constraints considered. Given a set of $n$ data points lying in an $m$-dimensional space, a matrix $X\in \mathbb{R}^{m \times n}$ is built such that each data point corresponds to a column of $X$. 
The goal of a low-rank matrix approximation is to express each data point as a linear combination of a few basis vectors. In other words, one has to find a matrix $W \in \mathbb{R}^{m \times r}$ and a matrix $H \in \mathbb{R}^{r \times n}$ such that each data point can be approximated as 
\[
{X(:,j) \approx \sum_{k=1}^r W(:,k) H(k,j)} \text{ for } j=1,\dots,n, 
\] 
where $W(:,k)$ denotes the $k$-th column of $W$ and corresponds to the $k$-th basis element and $H(k,j)$ is the weight with which the $k$-th basis element appears in the $j$-th data point. More precisely, $H(:,j)$ is the representation of data point $X(:,j)$ in 
an $r$-dimensional linear subspace spanned by $W$. 
In matrix form, this approximation, also sometimes called factorization, is  written as $X \approx WH$.  

On the other hand, deep neural networks have been widely used as deep learning gained success in many supervised classification tasks~\cite{lecun2015deep,marcus2018deep} and even in generative models~\cite{goodfellow2014generative}. Their main advantage lies in their ability to combine features in a highly non-linear way but their theoretical foundations remain quite elusive.


At midway between the linear algebra models and the deep neural networks lies deep matrix factorization (deep MF), the core of this paper. The main motivation of deep MF is to combine both interpretability, as in classical matrix factorizations, of which it is an extension, and the extraction of multiple hierarchical features, as allowed by multilayer architectures. One layer matrix approximations are not able to extract multi-level features in complex data sets. The goal of deep MF is to decompose a data matrix $X\in \mathbb{R}^{m \times n}$ as 
\begin{equation}
 X \; \approx \; W_1 \, W_2 \, \cdots \, W_L \, H_L , 
 \label{eq:very_first} 
\end{equation}
 where $L$ is the number of layers, $W_l \in \mathbb{R}^{d_{l-1} \times d_l}$ for  $l=1, \cdots, L$ with $d_0=m$, and $H_L \in \mathbb{R_+}^{d_L \times n}$. 
 The approximation in~\eqref{eq:very_first} corresponds to successive factorizations of $X$:
 \begin{equation}  \label{eq:keyAlgo}
\begin{split}
  X &\approx W_1 H_1, \\
  H_1 &\approx W_2 H_2, \\
  &\vdotswithin{=} \\
  H_{L-1} &\approx W_L H_L,  
  \end{split}  
\end{equation} 
where $ H_l \in \mathbb{R_+}^{d_l \times n}$ for all $l$. Each matrix $W_l$ ($l=1,\dots,L$) can be interpreted as the feature matrix of layer $l$ and each $H_l$ can be interpreted as the representation matrix of layer $l$. In other words, successive factorizations of rank $d_l$ ($1\leq l \leq L$) are performed such that various recombinations of the features of the first layers would appear in the following ones allowing numerous interpretations of the semantics hidden in the data set. 

\begin{figure*}[t]
\centering
\includegraphics[scale=0.45]{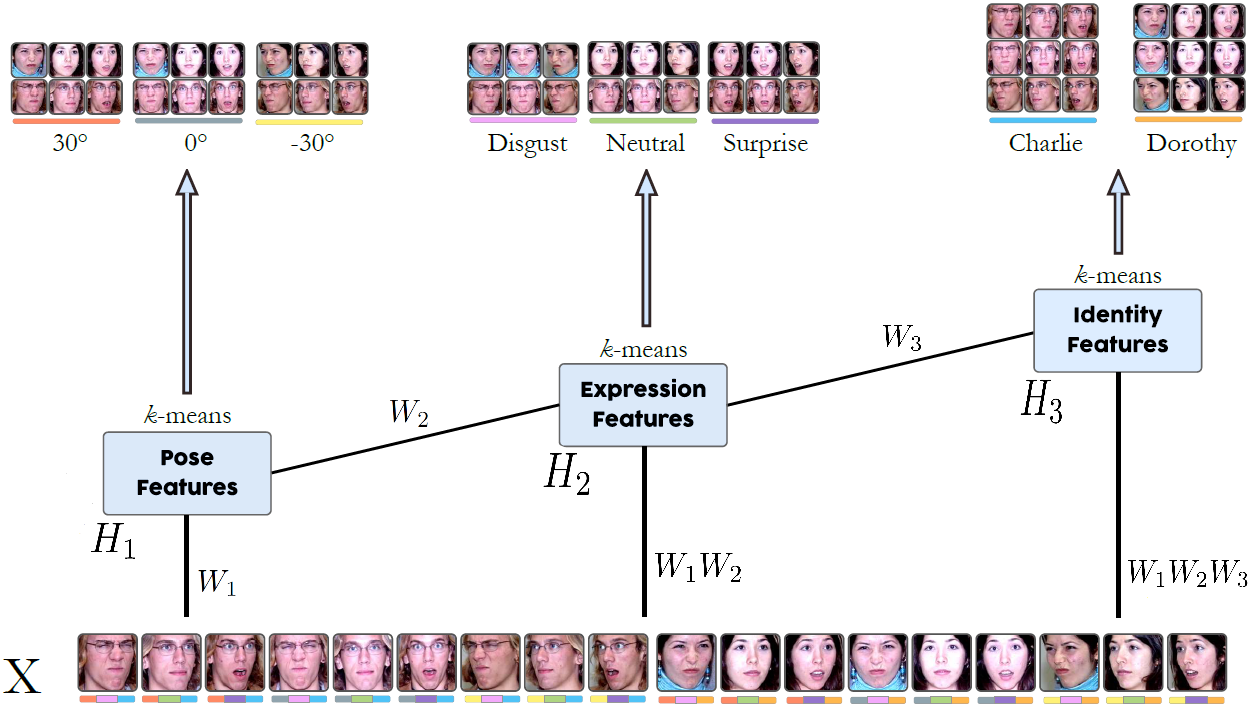}
\caption{Hierarchical features extracted by deep MF on the CMU-PIE face data set. 
At each layer $l$ ($l=1,2,3$), the columns of the representation matrix $H_l$ are clustered according to k-means to obtain the clusters shown above.  
Figure from~\cite{trigeorgis2016deep}.}
\label{fig:faces_trigeorg}
\end{figure*}

One of the first applications for which deep MF has proven to be useful is in the seminal paper of Trigeorgis et al.~\cite{trigeorgis2016deep} 
for the extraction of facial features. 
Given a set of $n$ gray-scale facial images, 
each one described by $m$ pixel values, deep MF extracts several layers of features, each one corresponding to a specific interpretation ranging from low-level features at the first layer to high-level features at the last layer. Fig.~\ref{fig:faces_trigeorg} illustrates such a decomposition for a factorization of depth $L = 3$ on the CMU-PIE face data set.  
The basis matrix $W_1$ contains $d_1$ archetypes of pose features, that is, $d_1$ faces having discriminative pose attributes, $W_1W_2$ corresponds to $d_2$ basis faces having different expressions, and $W_1W_2W_3$ retrieves the identities of the faces. 
Each $H_l$ indicates in which proportions each feature appears in each face of the original data set; for example the $j$-th column of $H_2$ contains the "proportions" in which the $j$-th subject is disgusted, surprised, or neutral.  
Note that the use of prior information, such as the fact that some faces share the same label at some layer, helps to improve the performance of the model; see Section~\ref{subsec:semisup} for more information.   

Without any constraint on the factors of deep MF,~\eqref{eq:very_first} merely degenerates into classical matrix factorization. In this case, the product of the matrices $W_l$'s could be replaced by a single equivalent (without additional particular property) matrix whose rank is less than or equal to the minimum of the $d_l$'s and the factorization is highly non-unique. 
For example, one could simply replace any $W_l$ by $W_lQ$ and $W_{l+1}$ by $Q^{-1} W_{l+1}$ for any $l$ and any invertible matrix $Q \in \mathbb{R}^{d_l \times d_l}$, and obtain another decomposition of $X$ with the same approximation error but most likely with a rather different interpretation. 
Therefore, constraints on the factors such as non-negativity and sparsity, and/or regularizations should be considered, which results in various deep MF models. 
Most deep MF models assume the non-negativity of several factors of the decomposition and therefore extend some NMF ideas. 

This paper serves as a survey on the recent literature on deep MF. It is organized as follows. 
We first briefly summarize the main ideas behind CLRMA in Section~\ref{secNMF}. 
In Section~\ref{secDeepMF}, we present the early multi-level models, namely multilayer MF up to recent deep MF, their regularizations, and the different algorithmic approaches to tackle them.  
In Section~\ref{secExp}, we present the performances of deep MF on two illustrative examples (namely, recommender systems and hyperspectral imaging), 
and review the main applications.  
Connections with deep learning are initiated in Section~\ref{subsec:neural} while Section~\ref{secTheo} highlights the lack of theoretical guarantees that have come with the models so far. However, contributions from deep linear networks might open new directions of research.  Finally, in Section~\ref{secPers}, we summarize the identified perspectives of future research and conclude.

\section{A brief summary on matrix factorizations}
\label{secNMF}

In this section, we recall the basics of matrix factorizations, which will be key to understand deep MF.

Low-rank matrix approximations consist in finding two matrices $W \in \mathbb{R}^{m \times r}$ and $H \in \mathbb{R}^{r \times n}$ such that the product $WH$ approximates as well as possible a data matrix $X \in \mathbb{R}^{m \times n}$ made of $n$ points in dimension $m$ where $r$, called the rank of the approximation, is generally a small value compared to $m$ and $n$ and is fixed in advance in many practical applications.

A critical aspect of matrix approximations is the choice of the loss metric between $X$ and $WH$, that is, the way to evaluate how good the approximation is. Most models aim at minimizing a divergence between the original data matrix and its low-rank reconstruction. More precisely, the $\beta$-divergences are usually considered to quantify the fidelity between the original data matrix and its low-rank approximation~\cite{fevotte2011algorithms}, and a common choice in the community is to minimize the Frobenius norm of the difference between these two matrices, which corresponds to $\beta=2$.
Therefore, the standard matrix factorization optimization problem is formulated as 
\begin{equation}
\label{eq:MF}
\underset{\substack{W \in \mathbb{R}^{m  \times r} \\ H \in \mathbb{R}^{r  \times n}}}{\min} \Vert X-WH \Vert_F^2,
\end{equation} 
where $||A||_F^2 = \sum_{i,j} A(i,j)^2$ is the squared Frobenius norm of matrix $A$. This essentially corresponds to PCA (although PCA typically performs mean centering before computing the principal components), and can be computed via the SVD. 

A usual feature of the data matrix is that it is entry-wise non-negative, that is, $X(i,j)\geq 0$ for all $i, j$, which is denoted $X \geq 0$.  Many real-world applications record such non-negative measurements, which has led to the development of the so-called non-negative matrix factorization (NMF) model \cite{wang2012nonnegative}. In NMF, the input data matrix $X$ is element-wise non-negative and in turn, entry-wise non-negativity is required for both factors $W$ and $H$. NMF has been widely studied, in terms of theoretical guarantees, models and applications~\cite{kim2014algorithms,fu2019nonnegative,gillis2014and,gillis2017introduction,cichocki2009nonnegative} and is formulated as  
\begin{equation}
\label{eq:NMF}
\underset{\substack{W \in \mathbb{R}^{m  \times r} \\ H \in \mathbb{R}^{r  \times n}}}{\min} \Vert X-WH \Vert_F^2 \; \text{  such that  }\; W \geq 0 \text{ and } H \geq 0,
\end{equation} which corresponds to~\eqref{eq:MF} with the additional non-negativity constraints.

A strong advantage of NMF is the interpretability of the factors~\cite{gillis2014and}. The matrix $W$ is often considered as the matrix of features, with each column of $W$ corresponding to a basis vector, while the matrix $H$ corresponds to the activations of each basis vector in each original data point. Especially, if it is also required that the sum of the elements of any column of $H$ is equal to $1$, that is, $H$ is column stochastic with  $\sum_{j=1}^r H(j,k)= 1$ for all  $k$, then 
the entries of the $k$-th column of $H$ can be interpreted as the proportions in which each feature vector appears in the $k$-th data point. 
 In this sense, NMF can be seen as a soft clustering technique as for all $j$ and $k$, $H(k,j)$ is the membership indicator of the $j$-th data point in the $k$-th cluster. 
 This model is sometimes referred to as simplex-structured matrix factorization; see~\cite{abdolali2020simplex} and the references therein. 

Most of the time, additional properties are enforced for the two factors $W$ and $H$. This can be translated by hard-coded constraints or through a penalty term called regularizer added to the data fitting term in the objective function. Several models and algorithms have been designed, exploiting geometric or algebraic properties~\cite{gillis2017introduction}. Among the most widely used techniques, minimum-volume NMF (MinVolNMF)~\cite{miao2007endmember, ang2018volume, xiao2019uniq}, sparse NMF~\cite{hoyer2002non}, and variants of archetypal analysis (AA)~\cite{morup2012archetypal,de2019near,javadi2019nonnegative} have led to the best performances. For example, minimum-volume NMF aims at minimizing the volume delimited by the basis vectors, while sparse NMF imposes that the factors only contain a reduced number of non-zero entries. Moreover, these methods start to be supported by theoretical advances, such as identifiability results which provide conditions under which  the ground-truth matrices $W$ and $H$ are unique (up to trivial ambiguities such as permutation and scaling); see~\cite{xiao2019uniq}  and the references therein. 
Some of these variants will be detailed in Section \ref{subsec:variants} as they have been extended to the multi-layer case. 

NMF is an NP-hard non-convex problem~\cite{vavasis2009complexity} which is generally solved through an alternated scheme known as block-coordinate descent (BCD), as described in Algorithm~\ref{algo0}. This consists in alternatively optimizing one of the two factors of~\eqref{eq:NMF} while keeping the other fixed. Note that the corresponding subproblems are convex, namely they are nonnegative least squares problems which are efficiently solvable\footnote{It can be solved for example in Matlab via the function \texttt{lsqnonneg}.}.    

\setlength{\textfloatsep}{0pt}
    \begin{algorithm} [H]
        \caption{Two-block coordinate descent to solve NMF} \label{algo0}
     
 \begin{algorithmic}[1]
 \renewcommand{\algorithmicrequire}{\textbf{Input:}}
  \REQUIRE Nonnegative matrix X, rank $r$ of the factorization
   \renewcommand{\algorithmicrequire}
  {\textbf{Output:}}
  \REQUIRE Matrices $W$ and $H$ minimizing~\eqref{eq:NMF}

 \STATE Compute initial matrices $W^{(0)}$ and $H^{(0)}$, $t=1$

  \WHILE {\textit{stopping criterion} not met}
  \STATE $W^{(t)}=\text{update\_W}(X,W^{(t-1)},H^{(t-1)})$ \\
  \STATE $H^{(t)}=\text{update\_H}(X,W^{(t)},H^{(t-1)})$ \\
  \STATE $t=t+1$ \\
  \ENDWHILE
 \end{algorithmic}
 \end{algorithm}
\section{Deep MF models and algorithms} \label{secDeepMF}

%

Although CLRMA such as NMF with non-negativity constraints and SCA with sparsity constraints lead to  a compact and meaningful representation of the input data, they are limited by the shallowness of the representation. Such techniques are only able to extract a single layer of features, preventing to reveal hierarchical features. While the standard matrix factorization decomposes the data matrix in only two factors, deep MF, inspired by the success of deep learning, is able to extract several layers of features in a hierarchical way, giving new insights in a broad range of applications. 

Deep MF considers a product of matrices $W_l$'s ($l=1,\dots,L$) in place of a single matrix $W$ in the approximation; see~\eqref{eq:very_first}. 
As constraints on the factors of this decomposition are necessary to make the model meaningful (see Section~\ref{sec:intro}), 
the next sections present various models and algorithms of deep MF. 
We first present the evolution from the early multi-layers models to the recent deep models in Section~\ref{subsec:begin}. Then, in Section~\ref{subsec:variants}, we describe the main variants, which are  inspired from those of classical matrix factorizations. Section~\ref{subsec:alg} describes the possible algorithmic choices and briefly discusses the computational cost.

\subsection{From multilayer MF to deep MF} \label{subsec:begin}

The first model extending CLRMA to several levels is multilayer NMF proposed by Cichocki et al.\ in 2006~\cite{cichocki2006multilayer,cichocki2007multilayer}. 
Based on the hierarchical factorizations of a non-negative data matrix $X \in \mathbb{R}_{+}^{m \times n}$ as described by~\eqref{eq:keyAlgo}, multilayer MF decomposes $X$ in a sequential manner. At the first layer, a low-rank factorization of $X$ is computed such that $X \approx W_1H_1$. At the second layer, the matrix $H_1$ is factorized as $H_1 \approx W_2H_2$, and so on until $H_{L-1}$ is decomposed as 
$W_L H_L$; see  Algorithm~\ref{algo1}. Moreover, all the factors of the decomposition are constrained to be non-negative.   
\setlength{\textfloatsep}{0pt}
    \begin{algorithm} [H]
        \caption{Early multilayer NMF~\cite{cichocki2006multilayer}} \label{algo1} 
 \begin{algorithmic}[1]
 \renewcommand{\algorithmicrequire}{\textbf{Input:}}
  \REQUIRE Non-negative data matrix X, number of layers $L$, inner ranks $d_l$ 
 \renewcommand{\algorithmicrequire}{\textbf{Output:}}
  \REQUIRE Matrices 
  $W_1,\dots, W_L$ and  $H_1,\dots,H_L$

 \STATE $Y=X$

  \FOR {$l = 1, \dots, L$}
 \STATE $(W_l, H_l)$ = Algorithm \ref{algo0} ($Y$, $d_l$)
  
 \STATE $Y=H_l$ \\
  
    \ENDFOR
 
 \end{algorithmic}
 \end{algorithm}
 
However, the multilayer NMF of~\cite{cichocki2007multilayer} does not investigate much the hierarchical power of deep schemes as the decomposition is purely sequential. 
More precisely, Algorithm~\ref{algo1} consists in sequentially minimizing the reconstruction errors $\|H_{i-1}-W_i H_i\|_F^2$ for all $i=1,\dots,L$ with $H_0=X$, but it does not involve a global cost function. In other words, the error is minimized layer by layer as in~\eqref{eq:NMF}, but there is no retroaction of the last layers on the first ones through some backpropagation mechanism.  


\begin{figure}
  \centering
  \subfloat[][]{\includegraphics[scale=0.45]{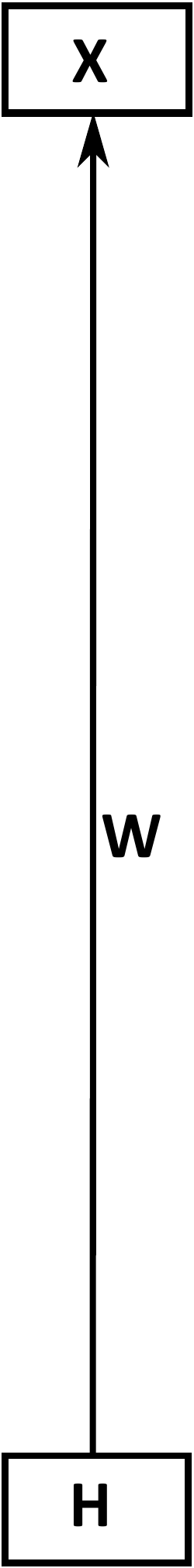}
  \label{fig:NMF}}
  \hspace{10mm}
   \subfloat[][]{\includegraphics[scale=0.45]{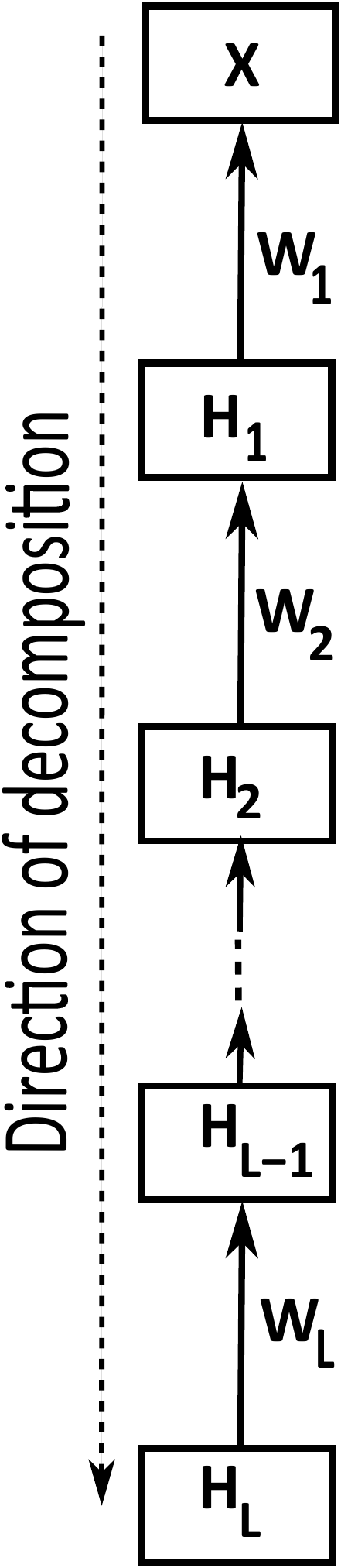}
  \label{fig:MlNMF}}
  \hspace{10mm}
   \subfloat[][]{\includegraphics[scale=0.45]{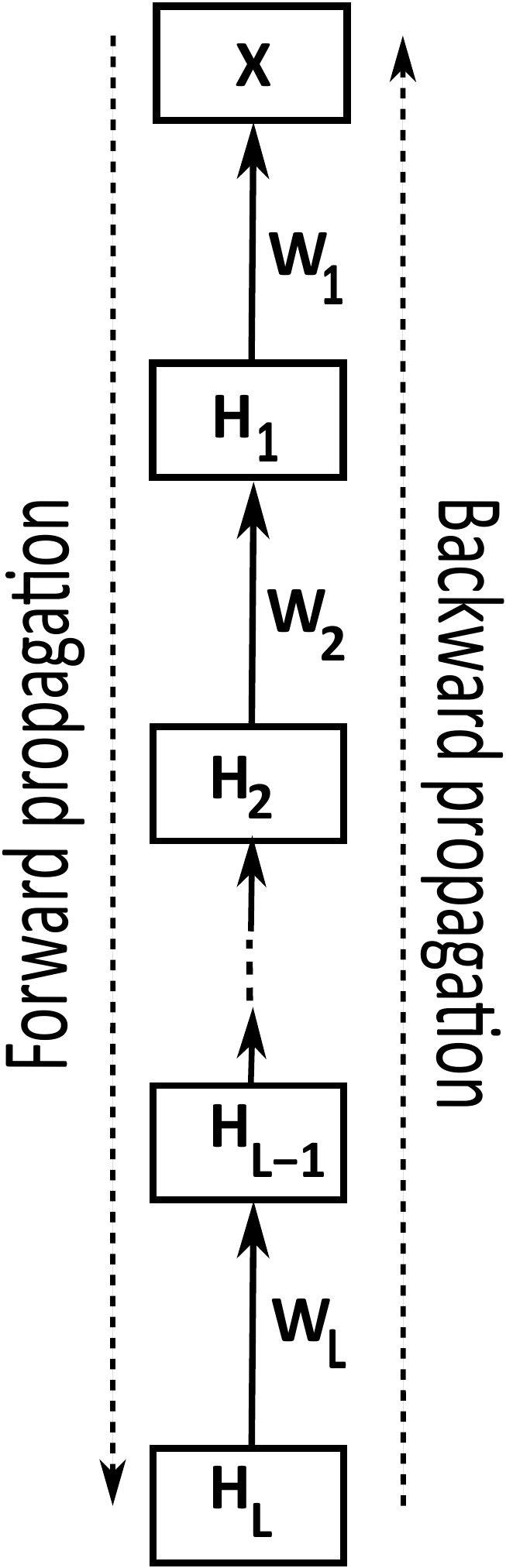}
  \label{fig:DeepMF}}    
\caption{ \protect\subref{fig:NMF} MF, \protect\subref{fig:MlNMF}  multilayer MF~\cite{cichocki2006multilayer}, \protect\subref{fig:DeepMF} deep MF~\cite{trigeorgis2016deep}. An arrow means a matrix product is performed: $\boxed{H} \hspace{-0.145cm} \xrightarrow[W]{} \hspace{-0.135cm} \boxed{X}$ means that $H$ is multiplied by $W$ to approximate~$X$. 
}
\label{fig:test}
\end{figure}

A key improvement 
was achieved by the papers of Trigeorgis et al., who introduced deep MF~\cite{trigeorgis2014deep, trigeorgis2016deep}. The data matrix $X$ still undergoes successive factorizations as in~\eqref{eq:keyAlgo}, but the breakthrough lies in the way the optimization of the factors $W_l$'s and $H_l$'s is performed.  The main algorithmic novelty is the fact that rather than using a purely sequential approach, the factors are iteratively updated and the information not only propagates from the first, more abstract layer, to the last, more refined layer, but also in the reverse direction. 
The following error function involving the factors of all layers is considered 
\begin{equation} 
\mathcal{L} (W_1, W_2,\dots, W_L; H_L)= \|X-W_1 W_2 \dots W_LH_L\|_F^2, 
\label{loss_deep}
\end{equation}
and a block-coordinate descent strategy is used to iteratively update all the factors. 
The deep MF algorithm~\cite{trigeorgis2016deep} is described in Algorithm~\ref{algo2}, and illustrated on Fig.~\ref{fig:DeepMF}. In Algorithm~\ref{algo2}, \textit{arg reduce} means that the factor is updated through some algorithm (see Section \ref{subsec:alg}) that (typically) decreases the objective function for several iterations.  

\setlength{\textfloatsep}{0pt}
    \begin{algorithm} [H]
        \caption{Deep  semi-NMF~\cite{trigeorgis2016deep}} \label{algo2}
     
 \begin{algorithmic}[1]
 \renewcommand{\algorithmicrequire}{\textbf{Input:}}
  \REQUIRE Data matrix X, number of layers $L$, inner ranks $d_l$
 \renewcommand{\algorithmicrequire}{\textbf{Output:}}
  \REQUIRE Matrices 
  $W_1,\dots, W_L$ and  $H_1,\dots,H_L$

 \STATE Compute initial matrices $W_l^{(0)}$ and $H_l^{(0)}$ for all $l$ through a sequential decomposition of $X$ (for example Algorithm~\ref{algo1})

  \FOR {$k=1,\dots$}
  \FOR {$l = 1, \dots, L$}
 
  \STATE $A_l^{(k)}=\prod_{j < l} W_j^{(k)} $ \\
  \STATE $B_l^{(k)}=\left\{
    \begin{array}{ll}
        H_L^{(k-1)} & \mbox{if } l=L \\
        W_{l+1}^{(k-1)}H_{l+1}^{(k-1)}  & \mbox{otherwise}
    \end{array}
\right.$
  \STATE $W_l^{(k)}=\underset{W}\argreduce \|X-A_l^{(k)} W B_l^{(k)}\|_F^2$ \label{line1}\\
  \STATE $H_l^{(k)}=\underset{H \geq 0} \argreduce \|X-A_l^{(k)}W_l^{(k)}H\|_F^2$ \label{line2}\\
  \ENDFOR

    \ENDFOR
 
 \end{algorithmic}
 \end{algorithm}

Several comments can be formulated:  

\begin{itemize}
\item First, the work of Trigeorgis et al.\ was inspired by semi non-negative matrix factorization (semi-NMF)~\cite{ding2010convex}, a variant of NMF where only one factor, typically $H$, must contain non-negative entries while the other $W$ is allowed to contain mixed-sign elements. 
Therefore, this model should rather be called deep semi-NMF as the $W_l$'s are not directly constrained to be non-negative, and the matrix $X$ is not required to have non-negative entries. 
However, one should keep in mind that each factor 
$H_l \approx W_{l+1}H_{l+1}$ is required to be non-negative, which implies an implicit constraint on the $W_l$'s. 

In practice, the requirement for non-negativity constraints on the basis vectors $W_l$'s  depends on the application: as most physical systems record non-negative data, it often makes sense to impose the non-negativity of the basis vectors. 
Therefore, if non-negativity of the basis vectors is meaningful, one can easily modify the model by adding nonnegativity constraints on the $W_l$'s, and modifying line~\ref{line1} of Algorithm~\ref{algo2}. 

\item Second, to initialize all factors, a forward decomposition of the input matrix is employed, as done in Algorithm~\ref{algo1}. Once all the factors are initialized, the updates of all matrices as in Algorithm~\ref{algo2} are performed until some stopping criterion is met. 

\item Third, the attentive reader may have noticed that Algorithm~\ref{algo2} does not correspond to applying a BCD method on~\eqref{loss_deep} by optimizing the factors $(W_1,\dots,W_L,H_L)$ alternatively. 
In fact, the nonnegative matrices $H_l$ ($l=1, \cdots, L-1$) are intermedate variables that do not appear in~\eqref{loss_deep}.  
However, one needs to remember the underlying sequential decomposition of~\eqref{eq:keyAlgo}: as $H_{l} \approx W_{l+1}H_{l+1}$ ($l=1, \cdots, L-1$) are constrained to be non-negative, they have a dedicated update rule. However, this raises important research questions that have not been investigated much. 
In particular, other possibilities in the expression of $B_l^{(k)}$ are possible; for example, \cite{yu2018learning} considers  $B_l^{(k)}=(\prod_{j > l} W_j^{(k-1)}) H_L^{(k-1)}$ while simply setting $B_l^{(k)}=H_l^{(k-1)}$ also makes sense, without a clear motivation why one should be preferred over the other. Moreover, how does replacing the function to minimize at line~\ref{line2} by $\|H_{l-1}^{(k)}-W_l^{(k)}H\|$ change the iterates $H_l$'s? 
If non-negativity constraints are imposed on the $W_l$'s, a classical BCD can be applied to optimize alternatively the factors of~\eqref{loss_deep}, as described in Algorithm~\ref{algo:BCD}, where only the variables $(W_1,\dots,W_L,H_L)$ are alternatively udpated. 
Note that nonnegativity constraints can be replaced with other constraints such as sparsity. 

\setlength{\textfloatsep}{0pt}
    \begin{algorithm} [H]
        \caption{BCD to minimize~\eqref{loss_deep}} \label{algo4}
     
 \begin{algorithmic}[1]
 \renewcommand{\algorithmicrequire}{\textbf{Input:}}
  \REQUIRE Data matrix X, number of layers $L$, inner ranks $d_l$
 \renewcommand{\algorithmicrequire}{\textbf{Output:}}
  \REQUIRE Matrices 
  $W_1,\dots, W_L$ and  $H_L$ minimizing~\eqref{loss_deep}

 \STATE Compute initial matrices $W_l^{(0)}$ for all $l$ and $H_L^{(0)}$

  \FOR {$k=1,\dots$}
  \FOR {$l = 1, \dots, L$}
   \STATE $A_l^{(k)}=\prod_{j < l} W_j^{(k)} $ \\
  \STATE $B_l^{(k)}=(\prod_{j > l} W_j^{(k-1)}) H_L^{(k-1)}$
  \STATE $W_l^{(k)}=\underset{W \geq 0} \argreduce \|X-A_l^{(k)} W B_l^{(k)}\|_F^2$ \label{line1}\\
 
  \ENDFOR
   \STATE $H_L^{(k)}=\underset{H  \geq 0} \argreduce \|X-W_1^{(k)}\cdots W_L^{(k)}H\|_F^2$ \\

    \ENDFOR

 \end{algorithmic}
  \label{algo:BCD}
 \end{algorithm}
 

\item Finally, the choice of the loss function itself is not obvious. In CLRMA, this issue has been investigated thoroughly, and several strategies exist, based on the statistic of the noise, or cross validation, among others~\cite{dikmen2014learning}. In deep MF, the question of the structure of the loss function also arises. 
Is a loss function of the type 
\[
D(X,W_1W_2 \dots W_LH_L ) , 
\]
where $D(A,B)$ is a similarity measure between two matrices $A$ and $B$, a good choice?  
Or would a loss function that balances the contribution of each layer, such as 
\begin{equation*}
\begin{split}
\mathcal{L}= D(X,W_1 H_1) + \lambda_1 D(H_1,W_2 H_2) + \dots \\ 
+ \lambda_{L-1} D(H_{L-1}, W_LH_L),  
\end{split}
\end{equation*}
be more appropriate? This question has not been addressed yet, to the best of our knowledge. 
Moreover, most works in the deep MF literature have only considered the Frobenius norm. 
It would be worth to investigate other similarity measures such as the Kullback-Leibler and  Itakura-Saito divergences, which have been shown to be particularly appropriate for specific applications in the case of standard NMF~\cite{fevotte2009nonnegative,leplat2020blind}.
\end{itemize}

To end up, a comparison of one-layer matrix factorization, multilayer MF~\cite{cichocki2006multilayer} and deep MF~\cite{trigeorgis2016deep} is illustrated on Fig.~\ref{fig:test}. Multilayer MF on Fig.~\ref{fig:MlNMF} and deep MF on Fig.~\ref{fig:DeepMF} both perform several levels of decomposition but the key difference is the iterative nature of the update rules in deep MF, while the decomposition is only sequential, that is,  unidirectional, in multilayer MF.

\subsection{Variants and regularizations of deep MF} 
\label{subsec:variants}

Beside the standard models presented in the previous section, some variants have been studied in the recent literature. 
These variants consist in adding constraints on the factors or enforcing some properties, and are mostly inspired from CLRMA. 
As highlighted in Section~\ref{sec:intro}, without any additional constraints on the factors, deep MF admits highly non-unique decompositions.  
The uniqueness of the solution is critical to ensure reproducibility and interpretability of the results. Depending on the application at hand, various constraints and regularizations can be used. 
In this section, we briefly review some of these models. In many of them, non-negativity is assumed on the factors, and the variants are therefore closely related to various NMF models.

\subsubsection{Deep orthogonal NMF}  
 \label{subsubsec:ortho}

Orthogonal NMF (ONMF)~\cite{ding2006orthogonal} is a variant of NMF which imposes that the matrix $H$ is nonnegative and row-wise orthogonal, that is, $H \geq 0$ and $HH^T=I_r$ where $I_r$ is the identity matrix of dimension~$r$. 
In other words, all rows of $H$ are orthogonal to each other, and their $l_2$ norm is equal to $1$. It is easy to see that these constraints imply that there is at most one non-zero value in each column of $H$. 
Hence each data point is only associated to one basis vector (one column of $W$), and ONMF is equivalent to a weighted variant of spherical $k$-means~\cite{pompili2014two}, which is a hard clustering problem. 
ONMF therefore imposes that each data point belongs to a single cluster which is represented by a single basis vector. 
This allows a very straightforward interpretation of ONMF factors: the columns of $W$ are cluster centroids, while the columns of $H$ assign each data point to its closest centroid (up to a scaling factor). 

A relaxation of the orthogonality constraint consists in adding a penalty term of the form $\sum_{j < k} H(j,:) H(k,:)^T$ to the objective function. This is referred to as approximately orthogonal 
NMF (AONMF)~\cite{li2014two}. 

The deep version of ONMF was introduced in~\cite{lyu2017deep} and enriched in~\cite{qiu2017deep}. The decomposition is slightly different than in multi-layer and deep MF because rather than having the activation matrices $H_l$'s successively decomposed, they decompose 
the features matrices $W_l$'s: 
\begin{equation} 
 \begin{split}
  X &\approx W_1 H_1, \\
  W_1 &\approx W_2 H_2, \\
  &\vdotswithin{=} \notag \\
  W_{L-1} &\approx W_L H_L,  
 \end{split}
 \label{eq:other_dec}
 \end{equation}  leading to  $X \approx W_L H_L \cdots H_1$, with each $H_l$ constrained to be nonnegative and row-wise orthogonal, 
 that is,  $H_l \geq 0$ and $H_lH_l^T=I_r$ for all $l$. Similarly, deep AONMF adds a penalty to the objective that minimizes the inner products $H_l(j,:) H_l(k,:)^T$, for all $j \neq k$ in each layer $l$. Applying the successive decompositions over the basis matrices $W_l$'s rather than the activation matrices $H_l$'s as in~\cite{trigeorgis2016deep} seems more natural: it allows to directly interpret the basis vectors of a given layer as combinations of the basis vectors of the next layer. 
For example, if the ranks $d_l$'s are chosen such that $d_L=d_{L-1}-1$, $d_{L-1}=d_{L-2}-1$,\dots, $d_2=d_1-1$, each layer will merge two clusters of the previous layer while keeping the others unchanged, and hence deep ONMF performs a hierarchical clustering. This will be illustrated on two showcase examples in Section~\ref{subsec:showc}. 
 

\subsubsection{Deep sparse MF} 

A very common constraint considered in CLRMA is the sparsity of some factors, referred to as SCA and closely related to dictionary learning (see Section~\ref{sec:relatedmodels}).  
It consists in limiting the number of non-zero elements of $W$ and/or $H$. 
Many papers have studied the case of one-layer sparse MF, 
see for example~\cite{hoyer2002non,  eggert2004sparse, kim2008sparse,  gribonval2015sparse,   
cohen2019nonnegative} 
among others. 
The goal of sparse MF is to render the factors more interpretable. 
In particular, the fact that each column of $H$ contains only a few non-zero entries means that each data point is associated with a few basis vectors.  

The extension of sparse NMF to the deep setting was proposed in~\cite{guo2019sparse}. Based on~\eqref{eq:keyAlgo}, a $\ell_1$ norm penalty is considered either on each column of the matrices $W_l$'s and/or on each column of the matrices $H_l$'s. 
Similarly to shallow sparse MF, the goal of sparse deep MF is to obtain sparse and easily interpretable factors at each layer. 
The subproblems w.r.t.\ the regularized factor can be efficiently solved for example through a proximal gradient descent method, such as the (fast) iterative shrinkage thresholding algorithm ((F)ISTA)~\cite{beck2009fast}. Note that a normalization of the other factor should be used to avoid a pathological case where the entries of the factor for which sparsity is promoted tend to zero while those of the other factor tend to infinity, because of the scaling degree of freedom in such decompositions ($W_l H_l = (\alpha W_l) (H_l / \alpha)$ for any $\alpha > 0$).  Furthermore, 
using the same regularization parameter for all columns of a factor at a given layer is discouraged. In practice, several regularization parameters can be tuned automatically to ensure balanced levels of sparsity~\cite{gillis2012sparse}. 
Another sparse framework, inspired by multilayer NMF \cite{cichocki2006multilayer}, consists in adding a regularizer based on the Dirichlet distribution on the columns of the factors~\cite{lyu2013algorithms}.

There exists many ways to impose sparsity on the factors, such as  $\ell_0$ norm~\cite{peharz2012sparse} or $\ell_{1/2}$ norm~\cite{qian2011hyperspectral} regularizations, among others. Inspired by deep learning, dropout could also enforce sparsity. Dropout~\cite{srivastava2014dropout} consists in randomly ``dropping'' some activations during the learning process to improve generalization. It has recently been employed for one layer NMF~\cite{he2019dropout}, and was shown to be equivalent to a deterministic low-rank regularizer~\cite{cavazza2018dropout}. It would be interesting to see to what extent dropout might regularize deep MF networks as well. Therefore, deep sparse MF has not yet been explored to its full extent.
%
%
%
%
%
%

\subsubsection{Deep non-smooth NMF} 

Non-smooth NMF (nsNMF)~\cite{pascual2006nonsmooth} consists in using a so-called smoothing matrix $S$ between $W$ and $H$ in NMF which has the form 
\[
S=(1-\theta) I_r + \frac{\theta}{r} ee^T , 
\]  
where $e$ is the vector of all ones of appropriate dimension. 
Note that  nsNMF reduces to NMF when $\theta=0$. 
The parameter $\theta \in [0,1)$ promotes the sparseness of both $W$ and $H$. Let us briefly explain why. 
We have $WS = (1-\theta) W + \theta \bar{w} e^T$ where $\bar{w}$ is the average of the columns of $W$, that is, 
$\bar{w} =  \frac{1}{r}We$, and $\bar{w}$ is denser than any column of $W$. 
When $\theta > 0$, $WS$ therefore moves the columns of $W$ towards $\bar{w}$, and $W$ is sparser than $\tilde{W}=WS$. 
 
 Deep nsNMF (dnsNMF)~\cite{yu2018learning} introduces a smoothing matrix at each layer of~\eqref{eq:keyAlgo}, with a common fixed parameter $\theta$, that is, $X \approx W_1S_1H_1$ and $H_{l-1} \approx W_lS_lH_l$ for all $l=2,\cdots, L$ with $S_l=(1-\theta) I_{d_l} + \frac{\theta}{d_l}ee^T$. Note that multilayer non-smooth NMF was already proposed in 2013 in~\cite{song2013hierarchical} to empirically show how a multilayer architecture, very similar to the one proposed by Cichocki et al.~\cite{cichocki2007multilayer}, is able to extract features in a hierarchical way in the context of text mining, but they did not use any backpropagation strategy.

 
\subsubsection{Deep Archetypal Analysis} \label{secDeepAA}

Archetypal analysis (AA)~\cite{morup2012archetypal}, also known as convex NMF~\cite{ding2010convex} is a variant of NMF in which the basis vectors are constrained to be convex combinations of the data points. In other words, in addition to the constraints stated in~\eqref{eq:NMF}, one should have $W=XA$ where $A \geq 0$ and $A^Te=e$. 
Intuitively, the basis vectors are restricted to lie in the convex hull of the data points, and can be interpreted as extremal points of the data set. Although the fitting error might be higher than in standard NMF, the closeness of the archetypes to the convex hull of the data confers them a better interpretability. 

The first proposal of deep AA was made in~\cite{sharma2018ase}, to the best of our knowledge, for the acoustic scene classification task. Given a data matrix $X$ composed of $n$ temporal frames characterized by an $m$-dimensional features vector, a discriminative representation is learnt through successive AA decompositions performed in a greedy forward way: 
\begin{equation}
\begin{split}
  X &\approx XA_1 H_1, \\
  H_1 &\approx H_1 A_2 H_2, \\
  &\vdotswithin{=} \notag \\
  H_{L-1} &\approx H_{L-1} A_L H_L. 
 \end{split}
\end{equation}
However, schemes including a backpropagation stage do not seem to have been tested yet for deep AA.


Finally, the non-determistic deep AA of Keller et al. ~\cite{keller2019deep} approximates the data points by samples drawn from a Gaussian distribution whose parameters are learnt through a deep encoding phase and is based on the deep variational information bottleneck framework~\cite{45903}. Since the model is probabilistic and non-linear, the spirit is quite different from the one of deep MF models presented previously. 

\smallskip

Closely related to AA, concept factorization (CF) consists in approximating the basis elements as linear combinations of the data points, the difference with AA lies in the absence of the sum-to-one constraints on both $A$ and $H$. 
Again, the first model containing several levels of decomposition was purely sequential~\cite{li2015multilayer}, and was outperformed by more recent approaches based on the deep MF algorithm; 
see for example~\cite{zhang2020deep}. We refer the reader to~\cite{zhang2020survey} for a comprehensive review of shallow and deep CF.

\subsubsection{Semi-supervised settings}
\label{subsec:semisup}

While the models presented so far were all unsupervised, some deep MF models are able to cope with available prior information in a semi-supervised fashion, such as deep weakly-supervised semi-NMF~\cite{trigeorgis2016deep}. To handle side information, a weighted graph is built at each layer, where the nodes are the data points and two nodes are connected by an edge if they share the same label. In the simplest case, the graph weights, denoted by the $n \times n$ symmetric matrix $G_l$ for the $l$-th layer, are binary,  
that is, $G_l(i,j)=1$ if $X(:,i)$ and $X(:,j)$ share the same label 
w.r.t.\ the features extracted at layer $l$. A smoothness regularization term is added to the loss function of~\eqref{loss_deep} with the form: 

\vspace{-2mm}
\small
\begin{equation}
\sum_{l=1}^L \lambda_l \sum_{\substack{j, k=1 \\ j \neq k}}^n \|H_l(:,j)-H_l(:,k)\|^2G_l(j,k)=\sum_{l=1}^L \lambda_l \Tr(H_lL_lH_l^T)
\label{semisupeq}
\end{equation}
\normalsize
where $\Tr(.)$ denotes the trace of a matrix, that is, the sum of its diagonal elements, and $L_l=D_l-G_l$ is the Laplacian matrix at layer $l$ with $D_l$ a diagonal matrix such that $D_l(j,j)=\sum_{k=1}^n G_l(j,k)$ for $j=1,\dots,n$. 
Intuitively,~\eqref{semisupeq} enforces the hidden representations $H_l(:,j)$ and $H_l(:,k)$ of data points $j$ and $k$ that share the same label at layer $l$ to be as close as possible. 

When the available information is such that each data point might be associated with several labels, 
a dual-hypergraph Laplacian is built to grasp richer underlying information and is such that an edge can connect any number of vertices~\cite{meng2019semi}. 

\subsubsection{Deep tensor decomposition} 

\begin{figure}
    \centering
    \includegraphics[scale=0.45]{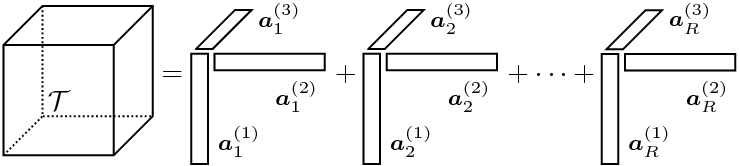}
    \caption{Canonical polyadic decomposition of a $3$-D tensor.}
    \label{fig:tensor}
\end{figure}

The extension of deep MF to tensors, that is, arrays of more than two dimensions, has not yet been much investigated. The analog of MF in the tensor world is the canonical polyadic decomposition (CPD), which decomposes a tensor as the sum of rank-one tensors; see~\cite{sidiropoulos2017tensor} and the references therein.  Given a tensor $\mathcal{T} \in \mathbb{R}^{I_1 \times I_2 \times \dots \times I_K}$ where $K$ is the dimension of the tensor, the CPD of rank $R$ aims at finding the vectors $a_j^{(i)} \in \mathbb{R}^{I_i}$ ($i=1,\dots,K$, $j=1,\dots,R$) such that 
\begin{equation}
\mathcal{T} \approx \sum_{j=1}^R a_j^{(1)} \circ  a_j^{(2)} \dots \circ a_j^{(K)}, 
\label{Tensor}
\end{equation}
where the $\circ$ operator denotes the outer product. 
The principle of a CPD is illustrated on Fig.~\ref{fig:tensor} for a $3$-dimensional tensor.

Multilayer frameworks of tensor decomposition have been proposed in~\cite{bi2018multilayer} for contextual-aware recommender systems (CARS), and in~\cite{casebeer2019deep} for audio sources separation in a mixed scene, as an extension of the matrix model of~\cite{smaragdis2017neural}. Deep tensor decompositions have also been introduced for action recognition in~\cite{jia2016sparse} and for the optimization of convolutional neural nets in~\cite{oymak2018end}. 

However, all these models are quite different from each other and do not resemble deep MF as defined in this paper. No general framework has been introduced for deep tensor decompositions, and it would be worth investigating deep CPD. 
A particularly interesting feature of CPD is that it has weak identifiability conditions (see for example~\cite{domanov2015generic} and the references therein), 
as opposed to standard MF, which could be leveraged in the deep setting. 

\subsubsection{Related models} \label{sec:relatedmodels}

In this section, we briefly introduce two models closely related to MF that have also been extended to a deep setting, namely transform learning and dictionary learning. There are not the main focus of the survey so we encourage the interested reader to look at the references for more details. 

Given an input data matrix $X \in \mathbb{R}^{m \times n}$, the goal of dictionary learning (DL) is to find a dictionary $D \in \mathbb{R}^{m \times r}$, whose columns are referred to as atoms, and a representation matrix $H \in \mathbb{R}^{r \times n}$ such that each column of $H$ is sparse, typically $k$-sparse (that is, with at most $k$ non-zero entries, where $k$ is a parameter). 
When the number $r$ of atoms is smaller than $m$, the dictionary is said to be undercomplete and DL is equivalent to SCA. 
However, DL typically looks for overcomplete dictionaries with $m \gg r$. 
A particularly interesting variant of DL is the convolutional sparse coding (CSC) where the atoms are convoluted with the representation matrix $H$. In~\cite{papyan2018theoretical}, Papyan et al.\ proposed deep CSC where successive factorizations are performed exactly as in~\eqref{eq:keyAlgo}, with sparsity constraints on the $H_l$'s; 
the columns of $H_l$ are required to be $k_l$-sparse, where $k_l$ is the desired level of sparsity at layer $l$. 
Similarly to deep MF, this decomposition allows a hierarchical interpretation of the dictionaries extracted. When a sequential thresholding algorithm is applied to enforce sparsity, deep CSC is equivalent to the forward pass of a convolutional neural network (CNN). 

Transform learning \cite{ravishankar2012learning} also shares similarity with MF. 
Given a matrix $X$, it consists in premultiplying $X$ with $W$ to obtain $H$, such that $WX \approx H$. The matrix $H$ is promoted to be sparse through a $\ell_1$ norm regularizer while $W$ is constrained to be full rank. Recently, a sequential multilayer transform learning framework was proposed by Maggu and  Majumdar~\cite{maggu2018unsupervised}, 
similar to the multilayer idea of Cichocki et al.~\cite{cichocki2006multilayer}. At the first layer, the approximation $W_1X \approx H_1$ is considered. Then, $H_1$ is  premultiplied such that $W_2H_1 \approx H_2$, and the process continues until $W_L$ and $H_L$ are found, with all $H_l$'s sparse and all $W_l$'s full-rank. Similarly to deep MF, the matrices $H_l$'s are hierarchical representations of the original data points that can be used for further clustering.

\subsection{How to solve deep MF?} \label{subsec:alg} 

In this part, we briefly describe the initialization techniques as well as the algorithms that can be used to solve the sub-problems w.r.t.\ either $W_l$ or $H_l$ at lines~\ref{line1} and~\ref{line2} of Algorithm~\ref{algo2}. As these algorithms are standard optimization techniques, we refer the reader to previous surveys~\cite{cichocki2009nonnegative,gillis2014and} for more details and references on their applications in CLRMA problems. We also discuss the choice of several parameters such as the number of layers $L$ and the inner ranks $d_l$'s.

\subsubsection{Initializations} 
The most commonly used initialization of deep MF consists in applying a sequential decomposition of the data matrix $X$, but there is no guarantee about the quality of this initialization. For the initialization of each factor $W_l$ and $H_l$ of each layer $l$, several routines have been presented in the literature; for example random initializations, initializations based on the SVD of $X$~\cite{trigeorgis2016deep}, or column subset selection methods that initialize $W$ with columns of $X$~\cite{gillis2014successive}. The study of initialization techniques dedicated to deep MF is still an open direction of research.

\subsubsection{Algorithms} Similarly to standard NMF, most algorithms for deep MF consist in alternatively updating each factor while keeping the others fixed as in Algorithm~\ref{algo2}. The stopping criterion can either be a maximum number of iterations, a sufficient decrease of the loss function, or a sufficient modification of the factors between two consecutive iterations.


The subproblems w.r.t.\ either $W_l$ or $H_l$ for any $l$ are typically  solved using standard first-order optimization algorithms. Trigeorgis et al.~\cite{trigeorgis2016deep} used a closed-form expression for the $W_l$'s and a multiplicative update (MU) for the $H_l$'s. MU is a well-known algorithm to solve NMF~\cite{lee2001algorithms}, and was also proposed to solve a sequential multilayer MF in~\cite{ahn2004multiplicative}. Other techniques such as projected gradient descent (PGD) method~\cite{lin2007projected}, possibly combined with an acceleration scheme, such as Nesterov's one~\cite{nesterov1983method} are widely used; see for example~\cite{guo2019sparse,yu2018learning}. PGD, a well-known first-order method to solve constrained optimization problems, is an extension of gradient descent (GD) where the iterates are projected on the feasible set at each iteration. 
The acceleration consists in adding a momentum term to the gradient step to allow faster convergence. Another standard optimization scheme is the alternating direction method of multipliers (ADMM) which consists in reformulating the problem by decoupling the variables, and minimizing the  augmented Lagrangian. It is standard in the CLRMA literature~\cite{huang2016flexible}, and it has also been used for  constrained deep MF in~\cite{zhou2018deep}. 

Since most deep MF algorithms are based on first-order methods, their computational cost is linear  w.r.t.\ the size of the input data and the ranks, 
and hence these methods are scalable. 
For example, the computational cost of the algorithm of 
Trigeorgis et al.~\cite{trigeorgis2016deep} requires $\mathcal{O}(Lt (
mnd +(m+n)d^2))$ operations, where $t$ is the number of iterations and $d=\underset{l=1,\dots,L} \max d_l$.

\subsubsection{Parameters}  \label{par:param}
The choice of the parameters of deep MF model, 
and in particular the number of layers and the inner ranks, mainly depends on the application. In most cases, the number of layers used for the results reported in the literature does not exceed three. 
Moreover, the ranks tend to be chosen in decreasing order as the first layers of the model are expected to capture attributes with a larger variance, thus requiring a larger capacity to encode them, while the last layers capture attributes with a lower variance~\cite{trigeorgis2016deep}. This observation is also derived from the analogy with autoencoders: the inner layer of an autoencoder is generally the one that contains fewer units as the goal is to obtain a compact representation of the input data; see Section~\ref{subsec:neural} for more details. 
On the other hand, when the decomposition is performed on the features matrices such as in~\eqref{eq:other_dec}, the ranks should also be chosen in decreasing order~\cite{qiu2017deep}: given $W_1 \in \mathbb{R}^{m \times d_1}$ with $d_1$ columns in dimension $m$, it only makes sense to approximate the columns of $W_1 \approx W_2H_2$ as linear combinations of $d_2 \leq d_1$ columns of $W_2$; see Section~\ref{subsec:showc} for two numerical examples.  




\section{Applications}  \label{secExp}

We now describe several applications for which deep MF is useful.

CLRMA has already been used successfully for countless real-world applications, and deep MF models have contributed to improve the performances. 
However, a clear definition of deep MF is absent within the community, 
and very diverse uses of this terminology have been used.  
Besides, \cite{arora2019implicit} mentions that some researchers call 
their model ``deep MF'' but use it for supervised tasks, or introduce a high degree of non-linearity inside it. 
This is for example the case of~\cite{fan2018matrix} that performs deep non-linear matrix completion, and~\cite{wang2017multi} where the inner representations are obtained through a deep highly non-linear MF architecture. 
The term ``deep MF'' was also given to an iterative procedure to solve classical MF through a deep unfolding of the iterations over time in~\cite{le2015deep}, 
though this has almost nothing in common with the deep MF as we have defined it in this paper.  

Therefore, in this part, we mainly focus on works in the same spirit as Trigeorgis~\cite{trigeorgis2016deep}, aiming to extract hierarchical features in a non-supervised context, which has led to breakthrough results in several applications. 
This section is organized as follows. In Section~\ref{subsec:showc}, we present two simple showcase examples showing the ability of deep ONMF to extract hierarchical features. 
We choose deep ONMF because, as explained in Section~\ref{subsubsec:ortho}, its factors are easily interpretable.  
A Matlab implementation is available~\footnote{\url{http://bit.ly/deepMF_v1}} to allow the interested reader to explore these deep MF examples, and play with the different parameters. 
Then, in Section~\ref{subsec:oth}, we present several applications for which deep MF has been successfully used in the recent literature.

\subsection{Two showcase examples}
\label{subsec:showc}

In this section, we detail two showcase examples on which deep MF reveals its inner workings. 
The first one is recommender systems (Section~\ref{subsubsec:rec}), and the second one is hyperspectral unmixing (HU) (Section~\ref{subsubsec:hyp}). 
For both applications, we describe the results obtained with deep ONMF (see Section~\ref{subsubsec:ortho}), which is a variant of deep MF particularly easy to interpret. Indeed, as each representation matrix $H_l$ contains only one non-zero entry per column, each data point is associated with a single cluster.

\subsubsection{Recommender systems} \label{subsubsec:rec}

Recommender systems consist in predicting the ratings of users over unseen items based on historical ratings on seen items. In other words, given an incomplete rating matrix $X \in \mathbb{R}^{m\times n}$ of $n$ users over $m$ items (such as movies), the goal is to predict the missing entries. 
 A standard approach to perform this task is by factorizing $X$ as the product of two matrices $W \in \mathbb{R}^{m \times r}$ and $H \in \mathbb{R}^{r \times n}$ where $W$ contains the ratings of $r$ basis users over the $m$ items and $H$ represents the proportions in which each user behaves 
 as the $r$ basis users~\cite{koren2009matrix}.  
 In the following, we describe how deep MF is able to extract hierarchical levels of basis users on a simple example. 

Let us consider a matrix $X \in \mathbb{R}^{9 \times 15}$ such that $X(i,j)$ contains the rating of user $j$ for movie $i$, and is between $1$ (highly dislike) and $10$ (highly like). Our goal is to apply deep MF on $X$ and show the hierarchy of basis users extracted. The matrix $X$ that will be considered throughout this synthetic example is the following: 
\smallskip

\hspace{-7mm}
\scalebox{0.93}{$
X= \begin{pmatrix}
7  &   8   & 10 &    8   &  9   &  4 &    1   &  2    & 3   &  2   &  4  &   1   &  3  &   5   &  2 \\
     8  &   8  &   7    & 8  &  10   &  5    & 1    & 2   &  6  &   1   &  4    & 2  &   2   &  1   &  2\\
     9  &   9   &  9   &  9    &10    & 4   &  1    & 4     &2  &   1   &  4     &3  &   1    & 1     &1\\
     3   &  1    & 2   &  1  &   3   &  8  &   8    & 7  &   8   & 10   &  3  &   4   &  1   &  2 &    2\\
     2   &  1  &   3  &   2   &  3 &    9   & 10    & 9 &    9 &    8  &   2   &  2 &    2  &   2 &    2\\
     1   &  2   &  1   &  1  &   2   &  8  &   8    & 9 &    9   &  8  &   3  &   2   &  3 &    3  &   1\\
     4   &  1   &  1   &  2   &  2   &  2   &  5    & 1    & 1    & 5   &  9  &   7    & 8   &  9  &   7\\
     2   &  2 &    2  &   1   &  2 &    2 &    6    & 2     &1   &  3    & 8 &    8  &   8  &   8   &  8\\
     1   &  2   &  2 &    1   &  3&     2  &   4    & 1  &   1    & 4   &  6   &  9   &  8   & 10   &  7
\end{pmatrix}.$}
\smallskip

Let us suppose, for example, that the first three  movies (first three  rows of $X$) 
are horror movies, 
the next three are comedy, and the last three  are biopics. 
We observe that the first five users mostly enjoy horror movies, the next five ones comedies, and the last five biopics. Note that we do not consider missing data in this example, as our goal is to interpret the deep MF decomposition, rather than predicting missing entries. 

We apply deep ONMF on $X$ 
with $L=2$, $d_1=4$, $d_2=3$, that is, by computing the following decomposition: 
\begin{equation} 
 \begin{split}
  X &\approx W_1 H_1, \quad H_1 H_1^T=I_4, \quad (W_1, H_1) \geq 0,\\
  W_1 &\approx W_2 H_2, \quad H_2 H_2^T=I_3, \quad (W_2, H_2) \geq 0.\\
   \end{split}
 \label{eq:other_decdec}
 \end{equation} 
To render the interpretation of the factors easier, we relax the orthogonality constraints by only imposing that $H_l H_l^T$ is diagonal, which does not change the hard clustering interpretation but simply allows each row of $H_l$'s to have a norm different from $1$. In counterpart, we normalize $W_l$'s such that all the elements are between $0$ and $10$. This allows an easier comparison between the features extracted at each layer. 

Let us interpret such a decomposition, layer by layer.  At the first layer, we have $X \approx W_1H_1$, and the matrices $W_1$ and $H_1$ are as follows (the values are rounded to one digit of accuracy):
\[
W_1= \begin{pmatrix}
 9.1  &  1.7   &     3.4 &3.8 \\
    8.9  &  1.1   & 4.9  &  2.7\\
   10.0 &   1.1  &   3.7  & 2.5\\
    2.2 &   10.0  &  8.6&   3.0 \\
    2.4 &  10.0  & 10.0 &  2.5 \\
    1.5 &   8.9  &   9.6 & 3.0 \\
    2.2 &   5.5  &    1.5 & 10.0\\
    2.0 &   5.0  & 1.8& 9.9   \\
    2.0  &  4.4   &  1.5 & 10.0\\
\end{pmatrix},
\]
\[
H_{1}= \begin{pmatrix}
0.88    &     0      &   0   &      0\\
    0.87   &     0    &     0        & 0\\
    0.92    &     0    &     0      &   0\\
    0.87    &     0     &    0     &    0\\
    1.05      &   0      &   0    &     0\\
         0   &      0    &0.92   &     0\\
         0  &  0.92       &  0  &       0\\
         0   &      0   & 0.85         &0\\
         0   &      0   & 0.92        & 0\\
         0   & 0.88      &   0       &  0\\
         0   &      0     &    0    &0.82\\
         0   &      0      &   0   & 0.80\\
         0   &      0       &  0  &  0.79\\
         0    &     0        & 0 &   0.89\\
         0    &     0         &0&    0.71
    \end{pmatrix}^T.
\] 
The columns of $W_1$ are themselves combinations of those of $W_2$, as $W_1=W_2H_2$ with $H_2$: 
\[
H_2= \begin{pmatrix}
1 & 0 &0 &0 \\
0 &1.02 &0.98 &0 \\
0 & 0 & 0 &1 
\end{pmatrix}.
\]
 
 At the second layer, we have $X \approx W_2 \hat{H}_2$, with $\hat{H}_2=H_2H_1$. 
As $d_2=3$, we expect that each column of $W_2$ corresponds to the profile of a basis user liking only one category of movies, which is indeed the case as we obtain  
\[
W_2= \begin{pmatrix}
9.1  &  2.7  &  3.8 \\
    8.9  &  3.3 &   2.7\\
   10.0  &  2.6 &   2.5\\
    2.2  &  9.1  &  3.0\\
    2.4  & 10.0 &   2.5\\
    1.5  &  9.3  &  3.0\\
    2.2  &  3.2  & 10.0\\
    2.0  &  3.1  &  9.9\\
    2.0  &  2.7  & 10.0
\end{pmatrix}, \;
\hat{H}_2^T = \begin{pmatrix}
0.88   &     0     &    0 \\
    0.87   &      0     &    0 \\
    0.92   &      0    &     0 \\
    0.87   &      0   &      0 \\
    1.05   &      0  &       0 \\
         0  &  0.91 &        0 \\
         0   & 0.94         &0 \\
         0   & 0.84        & 0 \\
         0   & 0.91       &  0 \\
         0   & 0.90      &   0 \\
         0    &     0   & 0.82 \\
         0     &    0    &0.80 \\
         0      &   0   & 0.79 \\
         0       &  0  &  0.89 \\
         0        & 0 &   0.71
\end{pmatrix}.
\] 
The first column of $W_2$ corresponds to a basis user liking horror movies, the second comedies, and the last biopics. 

The first and fourth columns of $W_1$ are identical to the first and third columns of $W_2$ respectively while the second and third column of $W_1$ bring more refined information. 
While the second column of $W_2$ only exhibits strong ratings for items $4$ to $6$ and low ones for the other items, the second and third columns of $W_{1}$ correspond to two different patterns such that the second basis user of layer $2$ can be seen as the merging of two more informative basis users at layer $1$. Both the second and third column of $W_1$ have high ratings over items $4$ to $6$, as for the second column of $W_2$ but the other ratings are different. Indeed, the second column of $W_1$ has intermediate ratings for biopics (items $7$ to $9$) but very poor ones over horror movies (items $1$ to $3$), and conversely for the third column. This level of granularity is not caught by the second layer of decomposition, which grasps the more general three main patterns that appear at first sight when looking at $X$. This justifies the benefit of using two layers of factorizations. To be more precise, as $d_1 > d_2$ in this example, deep MF first extracts $4$ refined basis users and then gather two of them at layer $2$ to model the more global structure of the data.

Let us mention that a single layer ONMF with $r=4$ recovers matrices similar to $W_1$ and $H_1$, and similarly when $r=3$ at the other layer. 
However, single layer factorizations do not exhibit any hierarchical relation between basis users.

\subsubsection{Hyperspectral unmixing} \label{subsubsec:hyp}

\begin{figure}
\centering 
\includegraphics[scale=0.55]{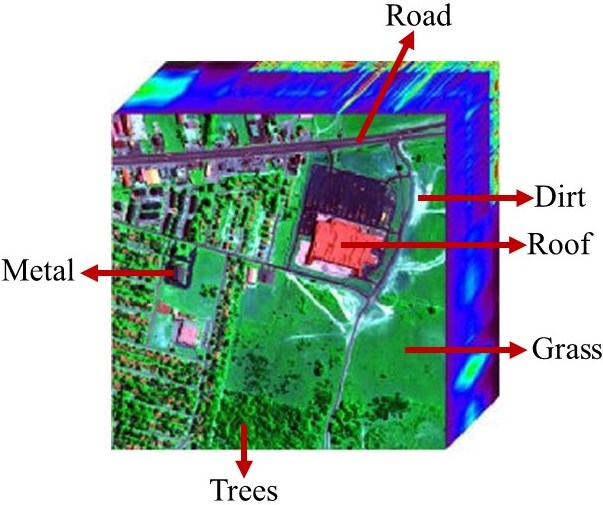}
\caption{The HYDICE Urban hyperspectral image. 
\label{fig:hyper_Urban}}
\end{figure}

Hyperspectral unmixing (HU) is a classical application of NMF, and many models taking into account various priors have been developed~\cite{bioucas2012hyperspectral, ma2013signal}. A hyperspectral image is composed of $n$ pixels, each one characterized by the reflectance value (fraction of the light reflected) in $m$ wavelengths, which is referred to as its spectral signature. Representing this image as a matrix $X \in \mathbb{R}^{m \times n}$ where each column is the spectral signature of a pixel, the purpose of HU is to identify the spectral signature of the $r$ materials present in the image, that is, the columns of $W$, as well as their respective abundances in every pixel, that is, the columns of $H$. However, the precise number of materials is not always easy to determine as some materials have similar spectral signatures or are highly mixed. 
In this section, we consider the HYDICE Urban hyperspectral image which is an airborne image of a Walmart in Copperas Cove (Texas); see Fig.~\ref{fig:hyper_Urban}. It is made of $n=307 \times 307$ pixels with $m=162$ spectral bands. There are several versions of the ground truth depending on the number of materials considered~\cite{zhu2017spectral}. 

\begin{figure*}
\hspace{-10mm}
      \includegraphics[scale=0.6]{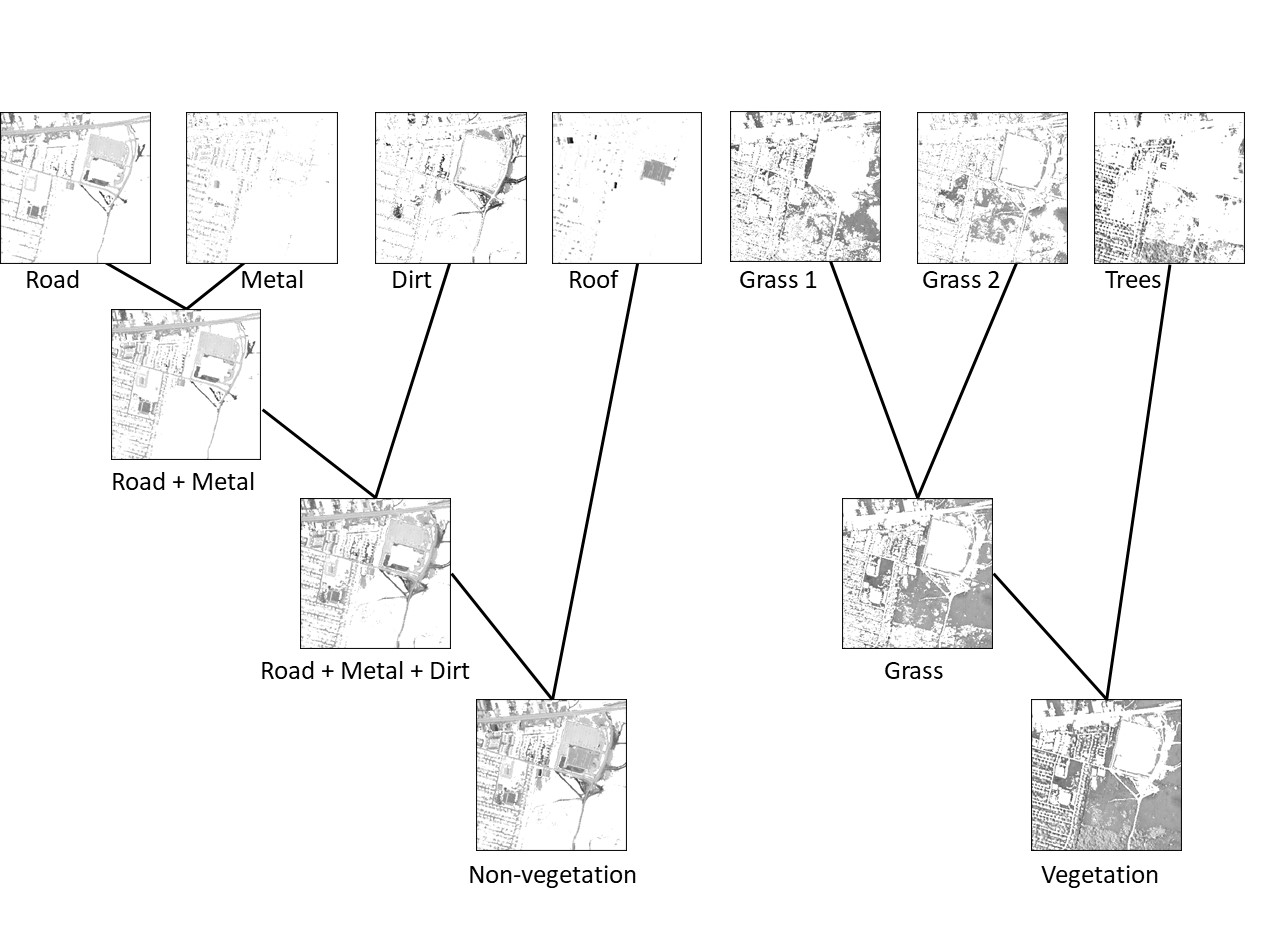}
\caption{Abundance maps hierarchically extracted by deep ONMF on the Urban data set. 
From top to bottom: first, second, third and fourth layer.} 
\label{fig:hyper_abund}
\end{figure*}

Deep ONMF is able to extract the materials in a hierarchical manner, 
as illustrated on Fig.~\ref{fig:hyper_abund} which shows the abundance maps $H_l$'s, representing the proportions of a given material in the pixels. 
This solution was obtained by applying deep ONMF on the Urban image with $L=4$,\; $d_1=7, \; d_2=6,\; d_3=4, \;d_4=2$.
The first layer extracts several materials, namely two types of grass, trees, road, dirt, metal and roof. 
At the next layers, the materials are successively merged by two within a single cluster.  
At layer $2$, the two clusters corresponding to road and metal, which have similar spectral signatures, are merged in a single cluster. 
At layer $3$, the road/metal and dirt are merged to create a single cluster while the two kinds of grass are also merged in a single cluster. 
At layer $4$, the road and roof are merged, while trees and grass are also merged in a cluster made of vegetation. Clearly, this example illustrates the ability of deep MF to extract materials in a hierarchical manner in hyperspectral images. Compared to traditional shallow extraction methods, deep MF brings an undeniable value in terms of interpretability.

\begin{figure*}
  \centering
   \subfloat[][]{\includegraphics[scale=0.48]{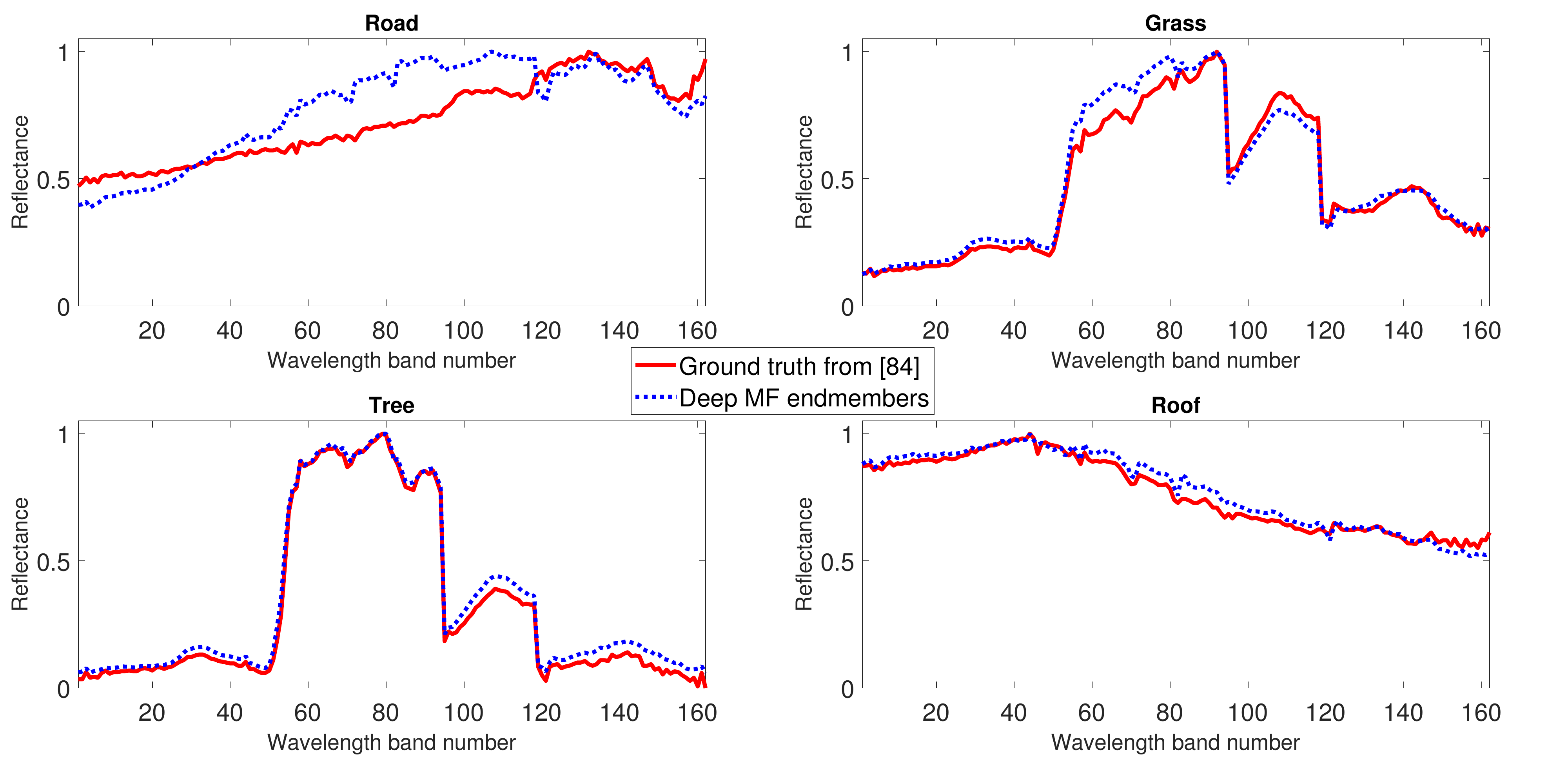}
  \label{fig:endmem_4}} 
\caption{Comparison between the endmembers extracted by deep ONMF at the third ($d_3=4$) layer and the "ground-truth" endmembers for the HYDICE Urban hyperspectral image. 
}
\label{fig:hyper_sign}
\end{figure*}

Fig.~\ref{fig:hyper_sign} provides a comparison between the extracted spectral signatures at the third ($d_3=4$) layer with a ground truth from \cite{Datarsla6:online}. 
The signatures retrieved by deep MF are similar to the ground truth, which indicates that deep MF is able to extract meaningful features through several layers. 


\subsection{Real-world applications of deep MF} \label{subsec:oth} 

In this section, we review several applications of deep MF presented in the literature.

\subsubsection{Recommender systems}

We have already shown that deep MF extracts hierarchical information in the context of recommender systems in Section~\ref{subsubsec:rec}. Several models based on deep MF have been proposed in the recent literature. 

Mongia et al.~\cite{mongia2020deep} use a projected gradient descent method to tackle deep MF with missing entries in the data matrix $X$. They refer to their model as deep latent factor model, and use it to infer missing entries while extracting several layers of explanatory factors, with a similar interpretation as in our showcase example in Section~\ref{subsec:showc}. 

Xue et al.~\cite{xue2017deep} derive a latent representation of both users and items through a so-called "deep factorization" though the model is different from the deep MF as defined in this paper. More precisely, based on a rating matrix $Y \in \mathbb{R}^{m\times n}$ containing both explicit ratings and non-preference implicit feedback (corresponding to a $0$ value) of $m$ users over $n$ items, each row $Y(i,:)$ is mapped to a vector $p_i$ such that 
$p_i=Y(i,:) W_1 \dots W_L$ and similarly (with another set of matrices) for each column $Y(:,j)$, which is mapped into $r_j$. Then, a matrix $\hat{Y}$ is built, such that $\hat{Y}(i,j) =\frac{p_i^Tr_j}{\|p_i\|\|r_j\|}$, and a cross-entropy based loss function between this similarity matrix and the original rating matrix $Y$ is minimized on a training subset of the data. This deep approach reaches higher performance than state-of-the-art MF methods in terms of the ranking of suggested items, based on their predicted ratings.

Deep MF was also used in the context of recommender systems with implicit feedback when the ratings are not given as a numerical value but as a binary feedback (such as "like" or "dislike")~\cite{yi2019deep}. For each user and each item, a vector containing both a representation of this implicit feedback and side information is provided. Then, deep MF is applied separately on the users and items to derive meaningful representations $H_L^{(u_i)}$'s and $H_L^{(v_j)}$'s for all users $u_i$'s and items $v_j$'s respectively, where the inner dimension $d_L$ is the same for both factorizations. 
The rating $r_{ij}$ of user $i$ over item $j$ is predicted as $r_{ij}=H_L^{{(u_i)}^T} H_L^{(v_j)}+G_{u_i} +G_{v_j}$ where $G_{u_i}$ and $G_{v_j}$ are obtained through maximum likelihood estimation and describe the specific influence of user and item, respectively. Using deep MF improves the root mean square error between the predicted and actual ratings compared to standard MF models on several benchmark datasets.

\subsubsection{Multi-view clustering}

Multi-view clustering consists in clustering items for which the data are described by several views;  for example, images described both by their pixels and textual tags, see~\cite{yang2018multi} for a survey. 
In~\cite{zhao2017multi} and~\cite{cui2019self}, each data matrix $X^{(v)}$ of each of the $V$ views is deeply factorized and the last hidden representations $H_L^{(v)}$'s are constrained to be the same for all views. Cui et al. design the objective function as a weighted sum of the squared Frobenius norm of the residuals of each view, whose weights are also learned in~\cite{cui2019self}. A further refinement is proposed by Wei et al. who add a penalty term measuring the redundancy of the clusterings $H_i$ and $H_j$ of different layers $i$ and $j$ to the objective function~\cite{wei2020multi}. More precisely, matrices $C^{(l)}=H_l^TH_l \in \mathbb{R}^{n \times n}$ for all $l$ indicate if two data points are clustered identically or not at layer $l$. A penalty aiming at minimizing $\|C^{(i)} \odot C^{(j)}\|_1$ is added to the objective for each pair of layers, to avoid redundancy, where~$\odot$~denotes an element-wise multiplication. 

A semi-supervised variant is considered by Xu et al.~\cite{xu2018deep} with a graph Laplacian penalty aiming to both minimize the gap between the inner representation $H_L$ of instances sharing the same label and maximize the gap between the inner representation $H_L$ of instances belonging to different classes. 
Huang et al.~\cite{huang2020auto} constrain the entries of $H_L$ to be either $0$ or $1$. 
Finally, when the data are given through several views, such as images and documents, Xiong et al.~\cite{xiong2020cross} show that binary hashing codes derived through deep MF are able to find meaningful items with a binary code close to the one of a given query.  
In addition to the data fitting error, the loss function contains terms that aim at finding a unified latent representation $H$, and at minimizing the classification error of a linear classifier based on $H$. Moreover, each entry of the unified code matrix $H$ is constrained to be either $+1$ or $-1$.

\subsubsection{Community detection}

Community detection consists in identifying communities, that is, subsets of nodes that are highly connected, inside a given graph. 
While NMF is able to extract overlapping communities~\cite{yang2013overlapping}, 
deep MF allows to interpret the dynamics along which the nodes are progressively grouped. 
More precisely, taking as input of deep MF the adjacency matrix leads to the extraction of the membership coefficients of all nodes to $d_l$ communities $H_l \in \mathbb{R}^{d_l \times n}$ at layer $l$ with $d_L \leq d_{L-1} \leq \cdots \leq d_0=n$~\cite{ye2018deep}. The interest of the deep architecture lies in the fact that nodes belonging to the same community gather closer to each other in terms of inner representations as the layers go deeper. 
In other words, deep MF allows to extract communities at different scales, smaller communities at the first layers are merged together in larger communities in the deeper layers as deep MF unfolds.

\subsubsection{Hyperspectral unmixing}

As illustrated in Section~\ref{subsec:showc}, deep MF can be used meaningfully for HU, extracting several layers of materials.  

The early sequential multilayer NMF of Cichocki et al.~\cite{cichocki2006multilayer} ~\eqref{eq:keyAlgo} was used, together with sparsity regularization, 
by Rajabi and Ghassemian~\cite{rajabi2014spectral}. Though the endmembers are estimated more accurately, no additional insight is given on the interpretability power of the model. 
Later, Tong et al.~\cite{tong2017hyperspectral} use 
the deep model of Trigeorgis et al.~\cite{trigeorgis2016deep} and show that it is efficient for the extraction of endmembers, though the interpretation of the successive inner representations is not emphasized. 
A similar approach takes into account an additional regularization~\cite{feng2018hyperspectral}: on the one hand a sparsity constraint is considered on the abundance matrix $H_L$ while on the other hand, a spatial regularization is applied through the total variation minimization (TVM).  
In a nutshell, TVM~\cite{rudin1992nonlinear} is a well-known regularization which consists in computing the differences between the abundances of each pair of adjacent pixels and minimizing their sum to reduce the noise and get a smooth abundance map. 
A deep purely sequential model of archetypal analysis (deep AA), similar to the one developed by~\cite{sharma2018ase} (see Section~\ref{subsec:variants}), was also used for HU in~\cite{zhao2016multilayer}.

\subsubsection{Synthetic aperture radar (SAR)}

SAR consists in analysing the changes that appear on the surface of the Earth through high-resolution images. Given two images of the same location at different times, the goal of SAR change detection is to produce a binary map indicating the changed and unchanged pixels over the considered period. 

Gao et al.~\cite{gao2017change} use deep semi-NMF to cluster the pixels in three categories: "unchanged", "changed" and "intermediate". Then, a more refined classification step accurately determines which pixels of the landscape have changed or not. 
Similarly, Li et al. propose to solve the SAR change detection problem with non-smooth deep MF~\cite{li2020deep}. The framework is again semi-supervised since a classification stage aims at reconstructing the label matrix based on the inner representation matrix $H_L$.

\subsubsection{Audio processing}

The audio source separation problem consists in extracting the frequential spectra of the sources contained in a sound recording as well as their respective activations over time. 
NMF has been shown to be efficient to solve this problem, when the matrix $X$ is a time-frequency representation of the input data, for example the spectrogram obtained with a short-time Fourier transform (STFT); see~\cite{fevotte2009nonnegative} and the references therein. 

Sharma et al.~\cite{sharma2017deep} use deep MF for speech recognition: given a matrix $X$ of $n$ frames, each one corresponding to a sequence of successive words, described by an $m$-dimensional vector corresponding to the well-known cepstral coefficients \cite{davis1980comparison}, a deep MF alternating sparse and dense layers factorizes $X$. The authors empirically notice that alternating sparse ($l$ odd) and dense ($l$ even) layers leads to more discriminative features, and the features used for the classification of the frames are obtained by applying a PCA on the concatenation of the inner representations $H_l$'s corresponding to sparse layers. \

Hsu et al.~\cite{hsu2015layered} apply deep MF on the spectrogram matrix of a set of spoken sentences to extract several layers of frequential basis features, and is better able to separate the speakers in a mixture than a simple one-layer NMF. Thakur et al.~\cite{thakur2018deep} used deep AA to extract sources based on the spectrograms of bioacoustics signals, with the dictionaries learnt at the first layers corresponding to archetypes on the convex hull of the data while deeper atoms being more in the center of the data. The classification accuracy obtained with a SVM based on the inner representations $H_L$'s is higher than other state-of-the-art classification methods. An extension was proposed in~\cite{thakur2019deep} with a more sophisticated classification approach.



%

%

\subsubsection{Perspectives} 
Deep MF does not seem to have been tested yet on several applications in which it has important potentialities. For example, in text mining tasks, it seems logical that hierarchical structures appear. For example, for NMF, given a word-by-document matrix $X$ where the entry $X(i,j)$ is the number of times the word $i$ appears in the document $j$, NMF allows to automatically extract topics as the columns of the basis matrix $W$, while $H$ indicates which document discusses which topic~\cite{lee1999learning}. In this context, deep NMF would be able to extract hierarchies of topics, from coarser to finer topics.  For example, the first layers would extract general topics such as politics, geography and sports, while the deeper layers would refine these topics in sub-topics. For example, sports would be divided into tennis, soccer and golf, while soccer would contain results from different competitions. 
Note that NMF is known to be a simple topic model equivalent to latent semantic analysis/indexing (PLSA/PLSI)~\cite{ding2008equivalence}, and designing refined deep models would be of particular interest, similarly as done for NMF~\cite{arora2013practical}. 

Though this survey focuses on linear deep MF, some applications would benefit from the introduction of non-linearities (see Section~\ref{subsec:neural} for more details). 
For example, in hyperspectral unmixing, scattering and various interactions may justify the use of non-linear models~\cite{dobigeon2013nonlinear}. Therefore, it would be interesting to investigate non-linear deep MF in view of the specific requirement of the applications. 

\section{Connections with neural networks}
\label{subsec:neural}

\begin{figure*}
\centering
\begin{tikzpicture}[x=1.5cm, y=1.5cm, >=stealth, scale=1.2, every node/.style={scale=0.9}]

\foreach \m/\l [count=\y] in {1,2,3,missing,4}
  \node [every neuron/.try, neuron \m/.try] (input-\m) at (0,2.5-\y) {};

\foreach \m [count=\y] in {1,missing,2}
  \node [every neuron/.try, neuron \m/.try ] (hidden1-\m) at (2,2-\y*1.25) {};

\foreach \m [count=\y] in {1,missing,2}
  \node [every neuron/.try, neuron \m/.try ] (hidden2-\m) at (5,2-\y*1.25) {};

\foreach \m [count=\y] in {1,missing,2}
  \node [every neuron/.try, neuron \m/.try ] (output-\m) at (7,1.5-\y) {};

\foreach \l [count=\i] in {1,2,3,m}
  \draw [<-] (input-\i) -- ++(-1,0)
    node [above, midway] {$\mathbf{X}(\l,:)$};

\foreach \l [count=\i from 1] in {1,s\textsubscript{1}}
  \node [] at (hidden1-\i.center) {g};
\foreach \l [count=\i from 1] in {1,s\textsubscript{1}}  
  \node [above] at (2.45,4.45-\i*3.45) {$\mathbf{M_1}(\l,:)$};

\foreach \l [count=\i from 1] in {1,s\textsubscript{P-2}}
  \node [] at (hidden2-\i.center) {g};
\foreach \l [count=\i from 1] in {1,s\textsubscript{P-2}}  
  \node [above] at (5.7,4.45-\i*3.45) {$\mathbf{M}_{P-2}(\l,:)$};

\foreach \l [count=\i from 1] in {1,2}
  \node [] at (output-\i.center) {g};

\foreach \l [count=\i] in {1,c}
  \draw [->] (output-\i) -- ++(1,0)
    node [above, midway] {$\mathbf{\tilde{Y}}(\l,:)$};

\foreach \i in {1,...,4}
  \foreach \j in {1,...,2}
    \draw [->] (input-\i) -- (hidden1-\j);

\foreach \i in {1,...,2}
  \foreach \j in {1,...,2}
    \draw [->] (hidden1-\i) -- (hidden2-\j);

\foreach \i in {1,...,2}
  \foreach \j in {1,...,2}
    \draw [->] (hidden2-\i) -- (output-\j);

\node [align=center, above] at (0,2) {Input\\layer $0$};
\node [align=center, above, scale=1.3] at (1,1.2) {$\mathbf{Z}_1$};
\node [align=center, above, scale=1.3] at (6,0.6) {$\mathbf{Z}_{p-1}$};
\node [align=center, above] at (2,2) {Hidden \\layer $1$};
\node [align=center, above] at (5,2) {Hidden \\layer $P-2$};
\node [align=center, above] at (7,2) {Output \\layer $P-1$};

\node[fill=white,scale=3,inner xsep=0pt,inner ysep=10mm] at ($(hidden1-1)!.5!(hidden2-2)$) {$\dots$};
\end{tikzpicture}
\caption{Illustration of an artificial neural network.}
\label{fig:neural_net}
\end{figure*}
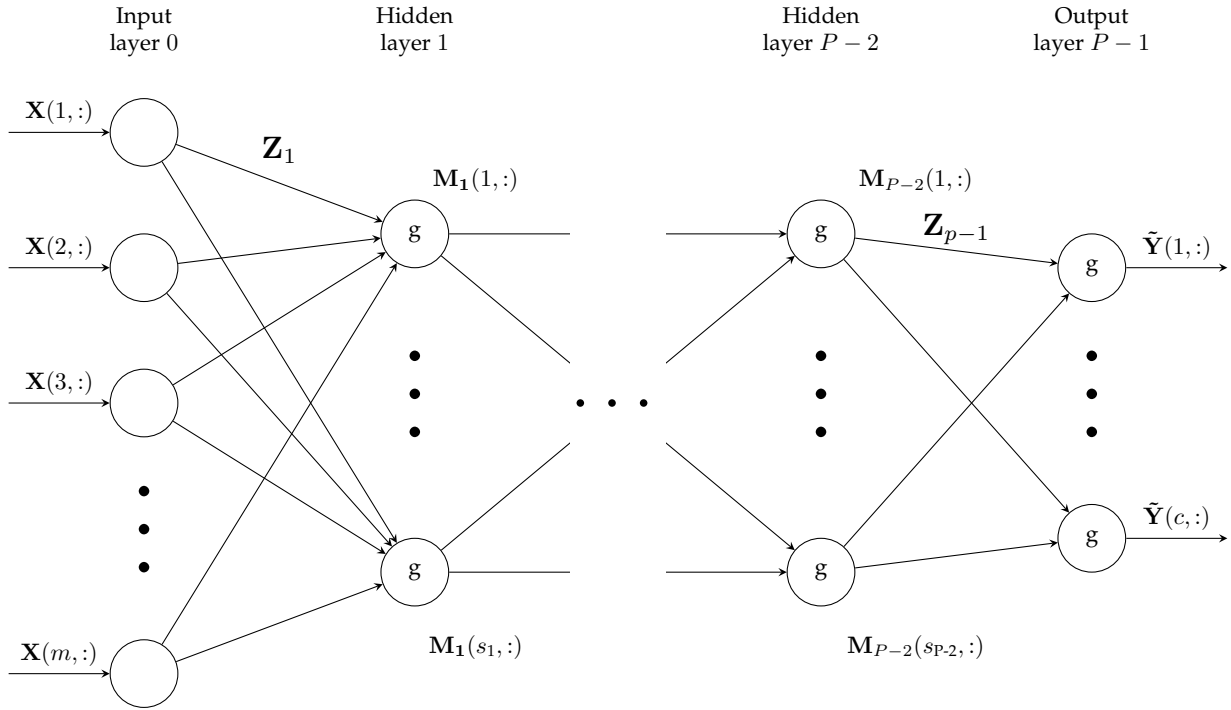

Several connections can be made between deep MF and deep learning. However, we restrict ourselves as much as possible to models aiming at extracting features from data in an unsupervised and interpretable way. Works embedding MF ideas in a neural network architecture, such as~\cite{sainath2013low,zhang2014extracting,kang2014nmf, ozkan2018endnet}  are interesting but are further away from the focus of this survey. 

Deep artificial neural networks~\cite{lecun2015deep} have been known for several years as one of the best classification paradigms. On Fig.~\ref{fig:neural_net}, we have represented a standard neural network made of a succession of $P$ fully-connected layers\footnote{We use an unusual naming of the parameters to avoid the confusion with the notation introduced for deep MF.}. 
Each layer $k$, $k=0, \dots, P-1$, is made of $s_k$ units. Let us consider a data matrix $X \in \mathbb{R}^{m \times n}$ of $n$ points in dimension $m=s_0$ and a binary label matrix $Y \in \mathbb{R}^{c \times n}$ indicating the membership of each data point $X(:,j)$ to each of the $c=s_{P-1}$ classes, that is, $Y(i,j)=1$ if $X(:,j)$ belongs to the $i$-th class. Given $X(:,j)$ for any $j$ as input, the network produces as output a $c$-dimensional vector $\hat{Y}(:,j)$. Calling $Z_k\in \mathbb{R}^{s_{k-1} \times s_k}$, $k=1,\dots, P-1$, the weights matrix between layer $k-1$ and layer $k$, the first layer computes a vector $M_1(:,j)=g(Z_1X(:,j))$ where $g$ is a non-linear activation function applied element-wise. Then, any layer $k$, $k=2, \dots, P-1$, computes $M_k(:,j)=Z_kM_{k-1}(:,j)$, with $M_{P-1}(:,j)=\hat{Y}(:,j)$. The goal of the neural network is to classify the data points $X(:,j)$'s at best, that is, optimize the $Z_k$'s such that the prediction $\hat{Y}(:,j)$ is as close as possible to the ground-truth $Y(:,j)$ for all $j$. Overall, considering all the data points, the prediction matrix is given by $\hat{Y} = g(Z_{P-1} g(Z_{P-2} \dots g(Z_1X)))$.

\tikzset{%
  every deepMF/.style={
    circle,
    draw,
    minimum size=0.8cm
  },
  deepMF missing/.style={
    draw=none, 
    scale=2.5,
    text height=0.07cm,
    execute at begin node=\color{black}$\dots$
  },
}

\tikzset{%
  every neuron/.style={
    circle,
    draw,
    minimum size=0.8cm
  },
  neuron missing/.style={
    draw=none, 
    scale=2.5,
    text height=0.07cm,
    execute at begin node=\color{black}$\dots$
  },
}

\begin{figure*}
\hspace{-4mm}
 \subfloat[][]{
 \begin{tikzpicture}[x=1.5cm, y=1.5cm, >=stealth, scale=0.88, every node/.style={scale=0.88}]
\begin{scope}[rotate=90]
\foreach \m/\l [count=\y] in {1,2,3,missing,4}
  \node [every neuron/.try, neuron \m/.try] (input-\m) at (-1,2.5-\y) {};

\foreach \m [count=\y] in {1,missing,2}
  \node [every neuron/.try, neuron \m/.try ] (hidden1-\m) at (0.2,2.5-1.5*\y) {};
  
\foreach \m [count=\y] in {1,missing,2}
  \node [every neuron/.try, neuron \m/.try ] (hiddenL_1-\m) at (2.1,1.35-0.9*\y) {};
 
\foreach \m [count=\y] in {1,missing,2}
  \node [every neuron/.try, neuron \m/.try ] (hiddenL-\m) at (3.3,0.75-0.6*\y) {};
  
  \foreach \m [count=\y] in {1,missing,2}
  \node [every neuron/.try, neuron \m/.try ] (hiddenL1-\m) at (4.6,1.35-0.9*\y) {};

\foreach \m [count=\y] in {1,missing,2}
  \node [every neuron/.try, neuron \m/.try ] (hiddenlast-\m) at (6.5,2.5-1.5*\y) {};

\foreach \m [count=\y] in {1,2,3,missing,4}
  \node [every neuron/.try, neuron \m/.try ] (output-\m) at (7.7,2.5-\y) {};

\foreach \l [count=\i] in {1,2,3,m}
  \draw [<-] (input-\i) -- ++(-1,0)
    node [left, midway] {$\mathbf{X}(\l,:)$};

\foreach \l [count=\i from 1] in {1,2}
  \node [] at (hidden1-\i.center) {g};
\foreach \l [count=\i from 1] in {1,s\textsubscript{1}}  
  \node [above] at (0.25,6-\i*4.35) {$\mathbf{M_1}(\l,:)$};

\foreach \l [count=\i from 1] in {1,2}
  \node [] at (hiddenL_1-\i.center) {g};
\foreach \l [count=\i from 1] in {1,s\textsubscript{Q-1}}  
  \node [above] at (2.2,4.65-\i*3.45) {$\mathbf{M}_{Q-1}(\l,:)$};
  
  \foreach \l [count=\i from 1] in {1,2}
  \node [] at (hiddenL-\i.center) {g};
\foreach \l [count=\i from 1] in {1,s\textsubscript{Q}}  
  \node [above] at (3.4,3.4-\i*2.6) {$\mathbf{M}_{Q}(\l,:)$};
  
  \foreach \l [count=\i from 1] in {1,2}
  \node [] at (hiddenL1-\i.center) {g};
\foreach \l [count=\i from 1] in {1,s\textsubscript{Q+1}}  
  \node [above] at (4.7,4.65-\i*3.45) {$\mathbf{M}_{Q+1}(\l,:)$};

\foreach \l [count=\i from 1] in {1,2}
  \node [] at (hiddenlast-\i.center) {g};
\foreach \l [count=\i from 1] in {1,s\textsubscript{P-2}}  
  \node [above] at (6.6,6.4-\i*4.65) {$\mathbf{M}_{P-2}(\l,:)$};

\foreach \l [count=\i from 1] in {1,2,3,m}
  \node [] at (output-\i.center) {g};
\foreach \l [count=\i] in {1,2,3,m}
  \draw [->] (output-\i) -- ++(1,0)
    node [left, midway] {$\mathbf{\tilde{Y}}(\l,:)$};

\foreach \i in {1,4}
  \foreach \j in {1,2}
    \draw [->] (input-\i) -- (hidden1-\j);

\foreach \i in {1,2}
  \foreach \j in {1,2}
    \draw [->] (hidden1-\i) -- (hiddenL_1-\j);
    
    \foreach \i in {1,2}
  \foreach \j in {1,2}
    \draw [->] (hiddenL_1-\i) -- (hiddenL-\j);
    
      \foreach \i in {1,2}
  \foreach \j in {1,2}
    \draw [->] (hiddenL-\i) -- (hiddenL1-\j);
    
       \foreach \i in {1,2}
  \foreach \j in {1,2}
    \draw [->] (hiddenL1-\i) -- (hiddenlast-\j);

\foreach \i in {1,2}
  \foreach \j in {1,4}
    \draw [->] (hiddenlast-\i) -- (output-\j);

\node [align=center, above] at (-1.3,3) {Input\\layer $0$};
\node [align=center, above, scale=1.3] at (-0.5,1.5) {$\mathbf{Z}_1$};
\node [align=center, above, scale=1.3] at (6.9,1.7) {$\mathbf{Z}_1^T$};
\node [align=center, above, scale=1.3] at (2.6,0.7) {$\mathbf{Z}_{Q}$};
\node [align=center, above, scale=1.3] at (3.8,0.7) {$\mathbf{Z}_{Q}^T$};
\node [align=center, above] at (-0.1,3) {Hidden \\layer $1$};
\node [align=center, above] at (1.8,3) {Hidden \\layer $Q-1$};
\node [align=center, above] at (3,3) {Hidden \\layer $Q$};
\node [align=center, above] at (4.3,3) {Hidden \\layer $Q+1$};
\node [align=center, above] at (6.1,3) {Hidden \\layer $P-2$};
\node [align=center, above] at (7.4,3) {Output \\layer $P-1$};

\node[fill=white,scale=1.8,inner xsep=15mm,inner ysep=0mm] at (1.2,-0.45)  {$\vdots$};
\node[fill=white,scale=1.8,inner xsep=15mm,inner ysep=0mm] at (5.5,-0.45)  {$\vdots$};
\end{scope}

\end{tikzpicture}
  \label{fig:autoenc}} 
  \rulesep
   \subfloat[][]{\raisebox{15mm}{\begin{tikzpicture}[x=1.5cm, y=1.5cm, >=stealth, scale=0.88, every node/.style={scale=0.88}]
\begin{scope}[rotate=90]
%
%
 
\foreach \m [count=\y] in {1,missing,2}
  \node [every neuron/.try, neuron \m/.try ] (hiddenL-\m) at (2.85,0.75-0.6*\y) {};
  
 \foreach \m [count=\y] in {1,missing,2}
  \node [every neuron/.try, neuron \m/.try ] (hiddenL1-\m) at (4.15,1.35-0.9*\y) {};

\foreach \m [count=\y] in {1,missing,2}
  \node [every neuron/.try, neuron \m/.try ] (hiddenlast-\m) at (6.05,2.5-1.5*\y) {};

\foreach \m [count=\y] in {1,2,3,missing,4}
  \node [every neuron/.try, neuron \m/.try ] (output-\m) at (7.25,2.5-\y) {};

%
%
  
\foreach \l [count=\i from 1] in {1,d\textsubscript{L}}  
  \node [above] at (2.65,3.55-\i*2.7) {$\mathbf{H}_{L}(\l,:)$};
  
  \foreach \l [count=\i from 1] in {1,2}
  \node [] at (hiddenL1-\i.center) {}; 
\foreach \l [count=\i from 1] in {1,d\textsubscript{L-1}}  
  \node [above] at (3.9,4.9-\i*3.6) {$\mathbf{H}_{L-1}(\l,:)$};

\foreach \l [count=\i from 1] in {1,2}
  \node [] at (hiddenlast-\i.center) {}; 
\foreach \l [count=\i from 1] in {1,d\textsubscript{1}}  
  \node [above] at (5.85,6.15-\i*4.45) {$\mathbf{H}_{1}(\l,:)$};

\foreach \l [count=\i from 1] in {1,2,3,m}
  \node [] at (output-\i.center) {}; 
\foreach \l [count=\i] in {1,2,3,m}
  \draw [->] (output-\i) -- ++(1,0)
    node [left, midway] {$\mathbf{\tilde{Y}}(\l,:)$};

%
%
    
      \foreach \i in {1,2}
  \foreach \j in {1,2}
    \draw [->] (hiddenL-\i) -- (hiddenL1-\j);
    
       \foreach \i in {1,2}
  \foreach \j in {1,2}
    \draw [->] (hiddenL1-\i) -- (hiddenlast-\j);

\foreach \i in {1,2}
  \foreach \j in {1,4}
    \draw [->] (hiddenlast-\i) -- (output-\j);

\node [align=center, above, scale=1.3] at (6.4,1.6) {$\mathbf{W}_1$};
\node [align=center, above, scale=1.3] at (3.3,0.7) {$\mathbf{W}_{L}$};
\node [align=center, above] at (2.6,3) {Layer $L$};
\node [align=center, above] at (3.9,3) {Layer $L-1$};
\node [align=center, above] at (5.85,3) {Layer $1$};

\node[fill=white,scale=1.8,inner xsep=18mm,inner ysep=0mm] at (5,-0.45)  {$\vdots$};
\node[fill=white,scale=2,inner xsep=12mm,inner ysep=0mm] at (-1.3,-0.45) {};
\end{scope}

\end{tikzpicture}}
  \label{fig:deepMF_eq} } 
\caption{Illustration of the similarity between \protect\subref{fig:autoenc} deep autoencoders and \protect\subref{fig:endmem_4} deep MF. 
} 
\label{fig:tikz}
\end{figure*}
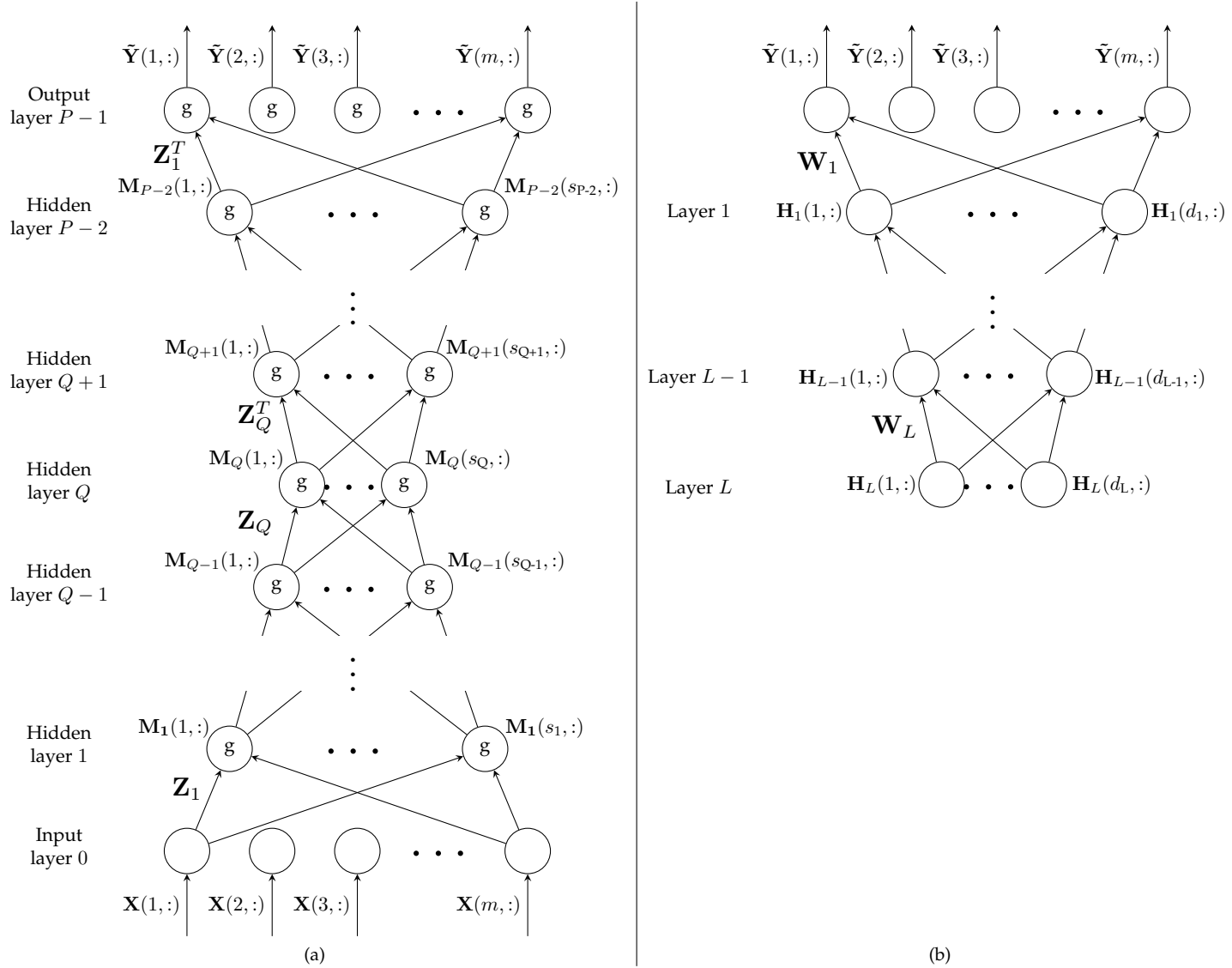

Autoencoders~\cite{ng2011sparse} are particular neural networks where the output matrix does not correspond to a membership matrix but is identical to the input, that is, $Y=X$. Assuming that the number of layers $P$ is odd, the purpose of an autoencoder is to extract a compressed representation $M_Q$ of the input data at the central layer $Q=\frac{P-1}{2}$ through the encoder, and approximate as well as possible the initial data back after the decoder layers. Fig.~\ref{fig:autoenc} provides an illustration when the encoder and decoder are symmetric, that is, $s_k=s_{P-1-k}$ for all $k=0,\dots,P-1$ and $Z_k=Z_{P-k}^T$ for all $k=1,\dots,P-1$. This leads to the following approximation 
\[
X\approx \tilde{Y}=g(Z_1^T g(Z_2^T \dots g(Z_Q^TM_Q))).
\] 

Let us number the layers in the reverse sense, that is, let us consider $l=P-1-k$ and $d_l=s_l$ for $l=1,\dots,L$. Let us also denote $W_l=Z_{P-l}=Z_l^T$ and $H_l=M_{P-1-l}$, with $L=Q$ such that the decoder performs the following decomposition:
\begin{equation}
X \approx \tilde{Y}=g(W_1 g(W_2 \cdots g(W_L H_L))).  
\label{eq:decoder}
\end{equation}
When the activation function $g$ is the identity,~\eqref{eq:decoder} becomes 
\begin{equation}
X\approx \tilde{Y}=W_1 \dots W_LH_L, 
\label{eq:lin_net}
\end{equation}
which corresponds to a so-called linear network. The decomposition performed by~\eqref{eq:lin_net} is the same as deep MF but deep MF usually requires additional constraints, such as the non-negativity of some factors, to render the solution meaningful and interpretable (see Section~\ref{sec:intro}).  


A widely used activation function is the rectified linear unit, that is, 
\[
g(x)=ReLu(x)=\max(x,0). 
\] 
In this setting, each inner representation matrix $H_{l-1}=g(W_lH_l)$ for $l=2,\dots,L$ is imposed to be non-negative, as in the original deep MF of~\cite{trigeorgis2016deep}. Though such a network is very similar to deep MF, as shown on Fig.~\ref{fig:deepMF_eq}, the two models are not exactly equivalent since the representation matrix $H_L$ of an autoencoder is learnt in a supervised way and is given by $H_L=g(W_L^Tg(W_{L-1}^T \dots g(W_1^TX))))$, which makes it close to deep archetypal analysis. In fact, autoencoders are mainly used in semi-supervised settings, for example to pre-train the networks for classification tasks, while deep MF mines unknown hierarchical features hidden in the data set. 
This connection suggests that the ranks of the factorization $d_l$'s in deep MF should be chosen in a decreasing order, as for the central layer of an autoencoder, corresponding to $H_L$, is usually the smaller one. 

Interestingly, the use of non-negativity constraints on the representation matrices within an autoencoder with a single layer in the encoder has shown to produce parts-based representations, as for NMF, while the overfitting was reduced~\cite{lemme2012online}. 
Also, improvements were achieved in~\cite{hosseini2016deep} using a deep structure promoting sparse activations $H_l$'s, which is similar to deep sparse MF. However, only the pretraining stage is unsupervised and is similar to deep MF (though non-linear activations are used), while a supervised classification stage follows. This connection between neural networks and deep MF was also highlighted in~\cite{flenner2017deep} where the discriminative power of such a hierarchical model was observed on topic mining and audio source separation tasks.

Also inspired by deep learning non-linearities, 
Trigeorgis et al.~\cite{trigeorgis2016deep} proposed to introduce a non-linear function, such as the sigmoid, at each layer of the model~\eqref{eq:keyAlgo}, that is, use  $H_{l-1}=g(W_lH_l)$ for all $l$ where $g(x)=\frac{1}{1+e^{-x}}$, which still provides a parts-based decomposition but at the cost of a possible weaker interpretability~\cite{tariyal2016deep}. 



Similarly, deep AA is closely related to neural networks. An archetypal regularization based on an autoencoder was proposed by van Dijk  et al.~\cite{van2019finding}. The latent representation $H$ is learnt through a deep encoder performing a non-linear transformation of the input data and the addition of Gaussian noise to $H$ enforces the basis vectors to be close to the data at the decoding layer. More precisely, the noise pushes the columns of $H$ outside the unit simplex, which in turn enforces the columns of $W$ to shrink in order to maintain a low reconstruction error.  The strong connection between autoencoders and deep AA in the process of learning hierarchical features in image patches was also highlighted in~\cite{bauckhage2015archetypal}.

%


 \section{Theoretical aspects of Deep MF} \label{secTheo}

Although numerous formulations, algorithms and application have been developed for deep MF, 
proper theoretical studies remain scarce, apart from insights from the deep learning community working on linear networks. To the best of our knowledge, the main theoretical contributions so far are mostly the convergence of algorithms, and to a lesser extent identifiability.

  \subsection{Convergence issues}
\label{subsec:conv}

When the factors of deep MF are updated through a BCD (see Algorithm~\ref{algo4}), the subproblems w.r.t.\ a single factor are convex, as for most CLRMA. 
Standard convergence results give conditions under which the iterates tend to stationary points, depending on whether the subproblems are solved through an exact algorithm or through an approximate framework, such as majorization minimization (MM). This encompasses most gradient descents used in practice. For example, when the global objective function is the least squares~\eqref{loss_deep}, using alternating projected gradient descent is guaranteed to converge to stationary points, because the subproblems are
convex and Lipschitz smooth~\cite{razaviyayn2013unified}. 

\subsubsection{Convergence of first-order methods} 

The effect of the number of layers on the convergence of first-order methods applied on the problem~\eqref{loss_deep}  has not been much studied, to the best of our knowledge, but recent results have been obtained on deep linear networks (see Section~\ref{subsec:neural}). 


Some of the theoretical results presented in the following are not directly related to the deep MF models described so far but rather concern networks aimed at supervised learning. However, we strongly believe that these insights coming from the deep linear networks community might be helpful to better understand deep MF and possibly open directions of future research. 
Let us mention a few important results. We refer the interested reader to the recent survey~\cite{sun2020global} for more details.    
\medskip

 When the thinnest layer of a deep linear network is either the input or the output one, Laurent et al.~\cite{laurent2018deep} showed that
 deep linear networks with arbitrary convex differentiable loss produce local minima that are all global. In addition, when the input data is whitened (that is, the covariance matrix is the identity) and a proper initialization of all layers is chosen, Arora et al.~\cite{arora2019convergence} proved the linear convergence of gradient descent to a global minimum on such a network. This generalizes the results of~\cite{bartlett2019gradient} in which linear residual networks, where the weights of each layer are initialized to be the identity matrix and the inner ranks $d_0=m,\dots,d_l$ are the same, are considered. Indeed, the network architecture is more general and softer restrictions on the initialization are required. 
When the loss function is the squared error between $Y$ and $\hat{Y}$, Arora et al.~\cite{arora2018optimization} provide an interesting result: If the weights $W_l$'s are updated with gradient descent and if the initialization $W_{l+1}^{(0)^T}W_{l+1}^{(0)}=W_{l}^{(0)^T}W_{l}^{(0)}$ holds for all $l$, then there exists an equivalent update rule for the end-to-end matrix $W=W_1 \cdots W_L$ which can be seen as an acceleration of the gradient descent update as long as the learning step is sufficiently small. As the depth $L$ grows, the effect is intensified which shows that over-parametrization, that is, considering several hidden layers, might accelerate the optimization process.

 
In the same spirit, when the network is restricted to be such that each hidden layer contains the same number $d$ of units, but without considering specific assumptions on the input data nor the initialization, Du et al.~\cite{du2019width} prove the linear convergence of gradient descent to a global optimum if the width of each layer is sufficient. 
  
Finally, the convergence of gradient descent on a function of a product of matrices, especially the loss $\|Y-\prod_l W_l\|_F^2$, where $Y=-I_d$ and each $W_l$ is a square matrix of size $d$ was studied by Shamir et al.\ in~\cite{shamir2019exponential}. Independent initialization of each layer is considered, that is either Xavier initialization (the entries of the $W_l$'s are sampled from a zero-mean Gaussian distribution) or near-identity initialization (each $W_l=I+M$ with $I$ the identity and $M$ a matrix whose elements are sampled from a zero-mean Gaussian distribution). The smaller the variance of the initialization distribution is, the more likely it is that gradient descent has exponential runtime w.r.t.\ the depth of the network, therefore advocating for shallow nets in this case. However, this result is rather empirical and the architecture of the network is particular, as each layer has the same number of units.

\subsubsection{Low-rank structure} 

Arora et al.~\cite{arora2019implicit} demonstrate some advantages of unconstrained deep MF compared to standard shallow MF~\cite{gunasekar2017implicit} in terms of regularization properties. Indeed, deep MF enhances an implicit tendency towards low-rank solutions. The problem considered is matrix completion, that is, impute missing entries in a given matrix $X \in \mathbb{R}^{m \times n}$. 
When the number of known entries is sufficiently large (this depends on the rank of $X$), factorizations of any depth admit solutions that tend to minimize the
nuclear norm of the end-to-end matrix $W=W_1 \cdots W_L$, that is, to minimize the sum of the  singular values of $W$. 
However, when there are fewer observed entries, the approximation tends to have a lower effective rank at the expense of a higher nuclear norm, especially when the depth increases. More interestingly, the evolution of the singular values of $W$ obtained with gradient flow, that is, gradient descent with infinitesimally small learning rate, reveals that the solutions tend to have a few large singular values and many small ones, with a gap that intensifies with the depth of the factorization. This can be seen as an implicit regularization promoting low-rank solutions. 

\bigskip

In summary, the recent literature gives evidence of the advantages of using a deep factorization, in terms of  both speed of convergence of gradient descent to a global minimum and  low-rank-\textit{ness} of the factors. However, the settings described do not assume constraints on the factors of the decomposition, unlike most deep MF models. Extending these observations to the constrained case is an important direction of research. 


\subsection{Identifiability}
\label{subsec:identif}

Identifiability of exact deep MF is an important theoretical research question. It consists in establishing the conditions under which the factors $W_1, \dots, W_L$ and $H_L$ can be uniquely retrieved, up to trivial permutation and scaling.  
For various CLRMA problems, thorough conditions have been proposed; 
see for example~\cite{fu2019nonnegative} for NMF, \cite{gribonval2015sparse} for SCA and DL, and \cite{abdolali2020simplex} for simplex-structured MF, and the references therein. 

Of course, any result for CLRMA can be extended to the corresponding deep MF model. 
Let us illustrate this with NMF. A necessary condition for an exact NMF $X = W_1H_1$ with $W_1 \geq 0$ and $H_1 \geq 0$ to be unique, up to permutation and scaling, is that $W_1^T$ and $H_1$ satisfy the so-called sufficiently scattered condition (SSC)~\cite{huang2013non}. Intuitively, the SSC requires that the rows of $W_1$ and the columns of $H_1$ are sufficiently well spread within the nonnegative orthant and have some degree of sparsity. 
Then, the two-layer $X=W_1 W_2 H_2$ is also unique, up to permutation and scaling, if $H_1 = W_2 H_2$ is unique, which is guaranteed if $W_2^T$ and $H_2$ satisfy the SSC. 
Similar observations would apply for SCA, among others. 

However, there are very few results tackling the identifiability of deep MF directly.  
As far as we know, the only attempt is by Malgouyres and Landsberg~\cite{malgouyres2016identifiability}, in a very particular setting. 
The factorization $X \approx M_1(q_1)M_2(q_2) \dots M_L(q_L)$ is considered where each matrix $M_l$ is described through a small number $S$ of parameters with $q_l\in \mathbb{R}^{S}$ for all $l$. A necessary and sufficient condition for identifiability in the noiseless case is provided as well as stability guarantees in the noisy case. However, these are quite abstract conditions involving advanced concepts such as the tensorial lifting property and the Segre embedding, and these conditions are difficult to check in practice. These results are further discussed in~\cite{malgouyres2019multilinear} where the conditions are extended to the case of convolutional linear networks. 


Needless to say that the robustness to noise (also sometimes referred to as the stability) of deep MF models is also an important issue that has not been investigated yet. In fact, even in the matrix case, most known results apply to the unconstrained case or under orthogonality constraints~\cite{stewart1990matrix}. For most other CLRMA problems, such results are rather scarce and difficult to derive.


 \section{Perspectives and conclusion} \label{secPers}


Deep MF is an emerging research topic, at the intersection of low-rank matrix approximations and deep learning. In this literature review, we presented multilayer and deep MF variants, which are being used successfully in an increasing number of applications, from recommender systems and hyperspectral unmixing to multi-view clustering and community detection.

Although many models and algorithms have been introduced for deep MF, 
the theoretical insights remain weak.
In our opinion, this is a main direction of research that should be tackled by deep MF researchers. Interestingly, a similar trend was observed for neural networks: the theory started to be investigated thoroughly (and it is still a very active area of research) only after many models 
and algorithms were shown to perform well in practice. 

Many perspectives have been presented throughout this survey, concerning the various aspects of deep MF:


\begin{itemize}
    \item \textbf{Choice of the parameters}: The choice of the parameters, namely the inner ranks and the number of layers, has not been discussed much as it is mostly application dependent (see Section \ref{subsec:alg}). 
Establishing proper guidelines to choose these parameters is a crucial issue. 
\item \textbf{Identifiability}: Identifiability of deep MF has not been investigated much, see Section~\ref{subsec:identif}. However, deriving conditions for deep MF to be unique could be particularly meaningful in some applications. 
\item \textbf{Loss function}: Very few works have carefully investigated the choice of the loss function, see the discussion in Section~\ref{subsec:begin}. Besides, this influences the way the algorithms are designed. 
\item \textbf{Design new models and algorithms}: As evoked in Section \ref{subsec:oth}, many MF models have not been extended to a deep setting yet. Moreover, efficient algorithms and initializations dedicated to deep MF are still lacking (see Section \ref{subsec:alg}). 
\item \textbf{Links between deep MF and deep learning}: Though there exist obvious connections between deep neural networks and deep MF as described in Section~\ref{subsec:neural}, they do not seem to have been fully exploited yet. It is not clear either whether it is possible to integrate advanced deep learning frameworks, such as convolutions, inside deep MF. 
Convolutional neural networks are known to extract several levels of visual features in image patches through highly non-linear operations~\cite{seddati2017towards}: is deep MF able of such performance in a linear and more interpretable way? 
\item \textbf{Applications}: Deep MF has not been applied yet on several important applications such as text mining. Moreover, in the applications described in Section \ref{subsec:oth}, the interpretation of the features obtained at each layer is not always clear. This is also an important research issue. In particular, the original data points are usually clustered by applying $k$-means on the last inner representation matrix $H_L$. However, more robust techniques taking into account the information of the previous layers have not been used yet, to the best of our knowledge.
\end{itemize}

\bigskip


We believe deep MF could be a particularly useful framework as it combines the ability to extract hierarchical features, as deep learning models, with a high interpretability power, as low-rank matrix approximations. 
These advantages justify the necessity to maintain research efforts in deep MF, especially to improve the explainability of AI techniques. 

 \medskip

\bibliographystyle{IEEEtran}
\bibliography{paper}

\end{document}